\documentclass[%
 preprint,
 amsmath,amssymb,
 aps,
 physrev,
 superscriptaddress,
]{revtex4-2}

\usepackage{graphicx}
\usepackage{dcolumn}
\usepackage{bm}
\usepackage{amsmath}

\DeclareMathOperator*{\argmin}{arg\,min}
\newcommand{\harpoon}{\overset{\rightharpoonup}}
\renewcommand{\vec}[1]{\mathbf{#1}}

\begin{document}

\title{Scientific Machine Learning of Chaotic Systems Learns Reduced-Order Equations for Neural Populations}

\author{Anthony G. Chesebro}
\affiliation{Department of Biomedical Engineering and Laufer Center for Physical and Quantitative Biology, State University of New York at Stony Brook, Stony Brook, NY, USA}
\affiliation{Athinoula A. Martinos Center for Biomedical Imaging, Massachusetts General Hospital and Harvard Medical School, Charlestown, MA, USA}

\author{David Hofmann}
\affiliation{Department of Biomedical Engineering and Laufer Center for Physical and Quantitative Biology, State University of New York at Stony Brook, Stony Brook, NY, USA}
\affiliation{Computer Science and Artificial Intelligence Laboratory, Massachusetts Institute of Technology, Cambridge, MA, USA}

\author{Vaibhav Dixit}
\affiliation{Computer Science and Artificial Intelligence Laboratory, Massachusetts Institute of Technology, Cambridge, MA, USA}

\author{Earl K. Miller}
\affiliation{Picower Institute for Learning and Memory, Massachusetts Institute of Technology, Cambridge, MA, USA}

\author{Richard H. Granger}
\affiliation{Department of Psychological and Brain Sciences, Dartmouth College, Hanover, NH, USA}

\author{Alan Edelman}
\affiliation{Computer Science and Artificial Intelligence Laboratory, Massachusetts Institute of Technology, Cambridge, MA, USA}

\author{Christopher V. Rackauckas}
\email{crackauc@mit.edu}
\thanks{Co-senior authors.}
\affiliation{Computer Science and Artificial Intelligence Laboratory, Massachusetts Institute of Technology, Cambridge, MA, USA}

\author{Lilianne R. Mujica-Parodi}
\email{mujica@lcneuro.org}
\thanks{Co-senior authors.}
\affiliation{Department of Biomedical Engineering and Laufer Center for Physical and Quantitative Biology, State University of New York at Stony Brook, Stony Brook, NY, USA}
\affiliation{Athinoula A. Martinos Center for Biomedical Imaging, Massachusetts General Hospital and Harvard Medical School, Charlestown, MA, USA}
\affiliation{Computer Science and Artificial Intelligence Laboratory, Massachusetts Institute of Technology, Cambridge, MA, USA}
\affiliation{Santa Fe Institute, Santa Fe, NM, USA}

\author{Helmut H. Strey}
\email{helmut.strey@stonybrook.edu}
\thanks{Co-senior authors.}
\affiliation{Department of Biomedical Engineering and Laufer Center for Physical and Quantitative Biology, State University of New York at Stony Brook, Stony Brook, NY, USA}
\affiliation{Athinoula A. Martinos Center for Biomedical Imaging, Massachusetts General Hospital and Harvard Medical School, Charlestown, MA, USA}
\affiliation{Computer Science and Artificial Intelligence Laboratory, Massachusetts Institute of Technology, Cambridge, MA, USA}

\date{\today}

\begin{abstract}
Extracting interpretable mathematical models from complex dynamical systems is difficult, especially for chaotic dynamics observed with noisy experimental data. We present PEM-UDE, a method that combines prediction-error methodology with universal differential equations to discover governing equations from limited, noise-corrupted observations. Prediction-error feedback smooths the chaotic optimization problem; for noise-free data generated within the model class, it preserves the data-consistent zero-loss set, whereas noise and model misspecification introduce a gain-dependent stability--bias trade-off. Preservation of the zero-loss set is not a guarantee of unique structural identifiability. We test the method on two benchmark chaotic systems---the R\"{o}ssler attractor and a real electrical circuit---and recover the correct functional forms even when one observed dimension contains noise of five times the signal magnitude. The method also accepts prior knowledge of the system as an initial functional form, which we use to learn neural circuit equations that account for sparse connectivity, a feature missing from conventional neural mass models. Applied to a population of Izhikevich neurons, PEM-UDE yields a multi-scale neural mass model that ties single-neuron parameters to macroscopic network dynamics and predicts a relationship between connection density, dominant oscillation frequency, and synchrony. We test these predictions against three intracranial recording datasets from rat and human cortices. For the neuroscience application, the learned equations are a reduced-order closure for a specified simulated Izhikevich network family; the experimental recordings provide an indirect consistency check of predicted frequency and synchrony trends, not a direct fit of the equations to recordings.
\end{abstract}

\maketitle

\section{Introduction}

Discovering the equations that govern complex systems has been an objective of dynamics for much of the past century, and typically requires both expert knowledge of the application domain and some understanding of statistical mechanics \cite{schmidt2009, king2009, cranmer2020, feinerman2018, strogatz_nonlinear_2015}. The approach has produced results across many fields, from climate models \cite{kochkov2024, foster2020} to chemical kinetics \cite{kim2021, ziepke2022}, human social interactions \cite{pagan2019}, and neural computation \cite{sejnowski_computational_1988}. Recent work has produced automated methods that assist the normally expert-driven process of describing data with systems of differential equations \cite{brunton2016, beregi2023, udrescu2020, ellis2023}. Symbolic regression techniques, including sparse identification methods (SINDy; \cite{brunton2016}) and genetic algorithms \cite{cranmer2023}, fit equations directly from data, but typically work less well with noisy data \cite{bakarji2023a}. Universal Differential Equations (UDEs) combine ideas from machine learning and scientific computing \cite{rackauckas2021}, embedding artificial neural networks in differential equation frameworks to learn unknown dynamics directly from data. The training of the network is constrained in two ways: by an imposed known functional form on the dynamics, with additional terms learned around it \cite{rackauckas2021}; and by structural constraints on the learning process itself \cite{ljung2002}. The contribution here is therefore specific: we do not introduce observers, UDEs, or symbolic regression as separate ideas, but combine observer-like prediction-error feedback with UDE training so that chaotic trajectories can be fit before the learned residual field is converted to an explicit symbolic expression.

These approaches struggle with chaotic systems, which are common in neuroscience and other biological contexts \cite{hannay2018, breakspear2017, chesebro2023, abrevaya2024effective}. Even simple physical systems can exhibit complex, chaotic solutions when certain parameters are brought past a bifurcation point \cite{rossler1976}, and neural dynamics often display characteristics of chaotic oscillators \cite{chesebro2023, schmidt2020}. The presence of chaos introduces numerical problems for the fitting procedure: the trajectory of a chaotic system is not well-defined beyond a Lyapunov time. Any numerical ODE solver introduces truncation error at each step, and sensitive dependence on inputs means that solver output will have an $\mathcal{O}(1)$ error in the produced trajectory shortly after a Lyapunov time. The only guarantee that the numerical simulation gives a meaningful trajectory is provided by the shadowing lemma, which states that there exists an $\epsilon$-ball around the initial condition in which the numerical solver's trajectory matches the trajectory of one of the other potential initial conditions within the $\epsilon$-ball, effectively describing the density of solutions around the attractor \cite{chandramoorthy2021}. The shadowing lemma has two consequences that make chaotic identification difficult. The divergent nature of chaotic trajectories creates instability in all sensitivity analysis methods used to compute gradients. Standard automatic differentiation through solvers then produces inaccurate derivative estimates unless specialized modifications are applied \cite{errico_what_1997, ni_adjoint_2019, ni_sensitivity_2019, chater_least_2017, wang_least_2014}. Consequently, the standard method for calibration of trajectories (fitting a loss function defined by numerical solutions of the ODE solver) is not a convergent algorithm in the context of chaotic systems. Noisy observational data can also produce a loss function that cannot distinguish between nearby trajectories during training, leading to inaccuracies in the fit dynamics \cite{chandramoorthy2021}. With exponential dependence of the solution on initial parameters, infinitesimal changes to the data may land the data point exponentially further along the time-series trajectory. Because of these properties, the long-term time series of a chaotic system can only be interpreted probabilistically, based on statistical or ergodic properties \cite{eckmann_ergodic_1985, petersen_ergodic_1989, bradley_nonlinear_2015}. The traditional technique of fitting a time series via the $L_2$ norm is therefore ill-defined in chaotic contexts. In real-world models, we additionally only have partial observability, which makes parameter fitting for any dynamical system difficult. This is particularly true in neuroscience, where neural interactions form complex systems operating across multiple scales: from molecular processes such as neurotransmitter activity to brain-wide electric fields that can be observed on EEG \cite{adam2024, breakspear2017, antal2024}.

Beyond the complexity inherent in describing these multi-scale interactions, brain dynamics are often poised near critical points that allow switching between different dynamic states depending on task requirements, metabolic resources, and the overall health of the neural circuits \cite{weistuch2021, habibollahi2023, tian2022}. The operation near critical points gives flexibility in information processing, but it also makes neural dynamics sensitive to small perturbations \cite{graf2024, chesebro2023}. Traditional approaches to modeling neuron populations, termed neural mass models, focused on capturing specific macroscale properties without a systematic approach to accounting for microscale biophysical parameters such as ion channel conductance \cite{wilson1972, jansen1995, larter1999}. Next-generation neural mass models (NGNMMs) have since provided an analytical expression for population-level activity that preserves the microscale detail present in individual neurons while also capturing emergent macroscale properties such as time-varying synchrony \cite{montbrio2015, chen2022, coombes2019, strogatz2000}. These analytical approaches typically rely on the Ott--Antonsen ansatz \cite{ott2008a} to avoid a Fokker--Planck derivation of the mean-field dynamics \cite{goldobin2024, divolo2018}. A caveat is that these derivations capture many desirable multi-scale dynamics but rely on non-biophysical assumptions, in particular all-to-all connectivity between neurons \cite{montbrio2015} and current conservation throughout the system \cite{divolo2018}. Recent work has extended these approaches to account for fluctuations arising from sparse connectivity and noise in networks of quadratic integrate-and-fire neurons \cite{goldobin2021}. Such extensions remain limited to simplified neuron models that lack biophysically detailed adaptation and recovery dynamics. Extending analytical mean-field derivations to biophysically detailed neuron models (e.g., Izhikevich neurons with adaptation currents), while at the same time relaxing connectivity assumptions, remains an open problem and motivates data-driven approaches.

The disconnect between the underlying neurobiology and the assumption of fully connected networks that grounds these derivations is consequential. The assumption is approximately true for deep and subcortical structures \cite{chen2022}, but far from physiological in the cortex, where only 1--5\% connectivity is typical \cite{pathak2024}. Cortical sparsity is not accidental: several lines of evidence indicate that the sparse, structured connectivity patterns in the cortex are essential for the structured information flow that supports higher cognitive processes \cite{balcioglu2023, fruengel2025}. Accounting for sparsity in equations that capture population activity is therefore needed for accurate models of the brain.

\begin{figure*}
    \centering
    \includegraphics[width=0.85\textwidth]{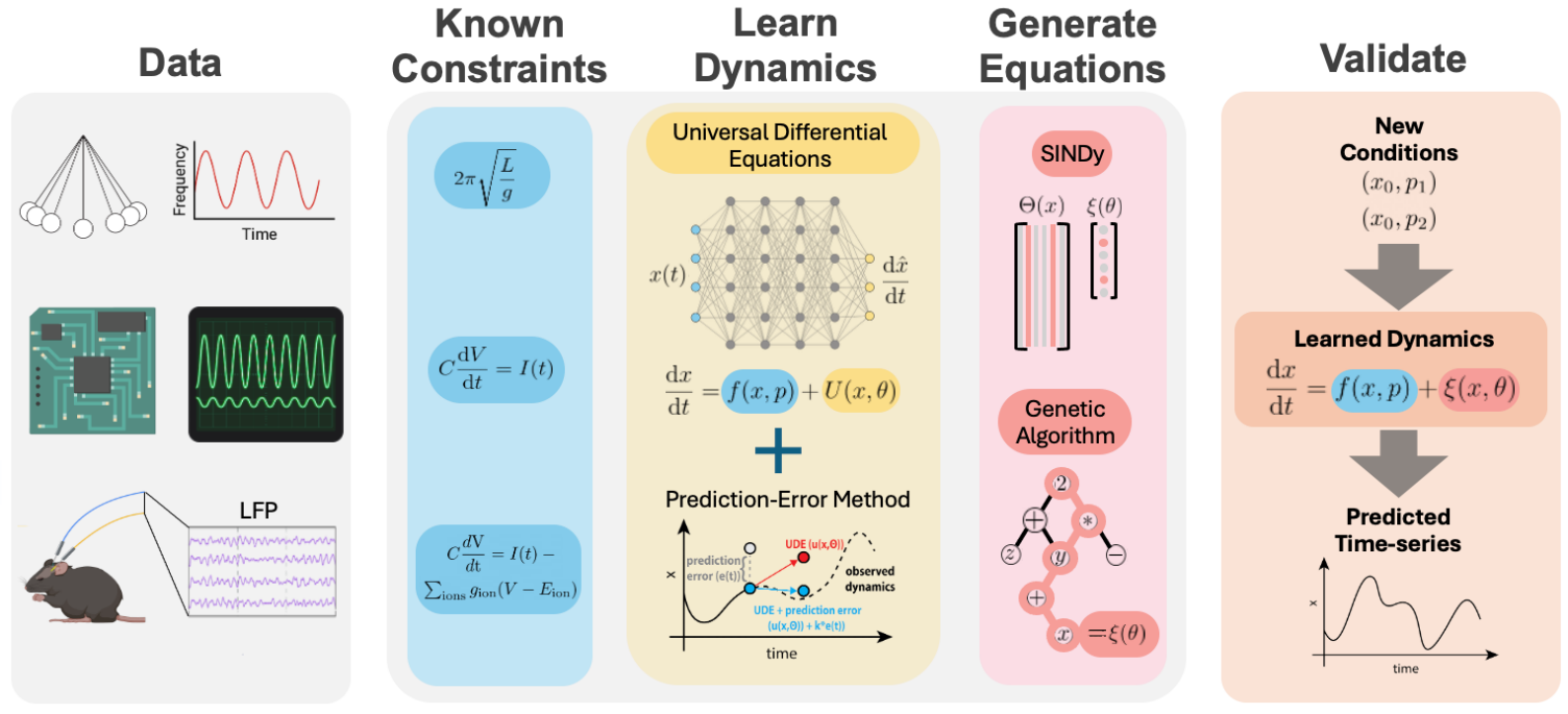}
    \caption{Schematic of the PEM-UDE approach. We begin with observations of a physical or biological system that have time-varying activity (data). From these, and from prior work, we form an initial description of the system under observation (known constraints). We then fit a UDE to describe the part of the dynamics not covered by this initial estimate, using the prediction-error method to help train the UDE on chaotic systems (learn dynamics). We then extract a symbolic form of the dynamics described by the UDE (generate equations) and check that this symbolic form correctly predicts activity beyond the initial observations (validate).}
    \label{fig:1}
\end{figure*}

To address these limitations, we present a method for training UDEs to fit chaotic system dynamics, and we use it to learn reduced-order closure terms for simulated neural population models across connectivity levels. We describe the procedure, which uses the prediction-error method (PEM) \cite{ljung2002, stoica1989} to train the UDE on chaotic system data, and then extracts a symbolic form of the learned expression from the UDE to generalize the dynamics. We show that this approach can learn unknown states in a chaotic system, accurately recovering a hidden state in the R\"{o}ssler system that traditional UDE approaches cannot, and that it does so by removing the sensitive dependence on parameter choices from the system. The approach also learns the dynamics of a chaotic electrical circuit under incomplete observation, where one dimension of the observed data is corrupted by noise of five times the magnitude of the original signal. Under these conditions, sequential thresholding least squares (STLSQ) applied directly to the data fails to recover the correct dynamic system, while PEM-UDE succeeds; we therefore argue that the method is more data-efficient than pure symbolic regression. Finally, from simulated firing-rate and voltage observations of a spiking network, we learn a reduced-order correction to an existing neural mass model. The symbolic correction generalizes across simulated connection probabilities and reveals how connectivity changes can alter the rhythmic dynamics of this network family. The later comparison with recordings tests whether these qualitative predictions match regional physiology; it does not claim direct equation discovery from the recordings. These controlled comparisons isolate the contribution of PEM relative to an otherwise identical UDE and direct STLSQ under the conditions studied; they are not a comprehensive benchmark of all methods for chaotic system identification. A graphical overview of the full pipeline is provided in Fig.~\ref{fig:1}.

\section{Results}

\subsection{Formal description of the PEM-UDE approach}

We give a formal definition of the combined PEM-UDE/symbolic regression method here, with implementation details in the Methods section and the accompanying code. The associated repository includes a demonstration using toy data that runs on standard computers, for readers who want to try the method out.

A graphical overview of the approach is provided in Fig.~\ref{fig:1}. We begin with observations of a physical or biological system that result in a dataset of related time series. The goal is a system of differential equations that describes the dynamics of the system under observation and accurately predicts future dynamics. From the observations and previous work on similar systems, we make an initial guess for the dynamics, using prior information from physical systems to structure the first approximation of the system's behavior \cite{rackauckas2021}. We then set up a UDE system to learn the unknown dynamics directly from the observed data. Standard methods for training UDEs struggle to fit chaotic systems accurately, as shown in the results below. To get around this, we developed a training approach that integrates PEM dynamics with the UDE during simulations, so that the UDE can learn from the data. Once training is complete, we extract a symbolic representation of the dynamics from the UDE network with either sparse linear solver-based symbolic regression (STLSQ) or evolutionary algorithm-based symbolic regression \cite{cranmer2023}, depending on the specific context and data availability. The symbolic description lets us both predict future time series and make analytical observations about the underlying dynamics. The prediction-error term is a training device: the symbolic equations are evaluated as autonomous dynamics, after the observer correction has been reduced or removed, so success requires the learned vector field itself to reproduce the observed dynamical structure.

\subsubsection{UDE formulation}

Following prior work \cite{rackauckas2021}, a universal differential equation with an additive correction of a general process $\vec{x}(t)$ with some observable state $\vec{y}(t)$ is given by
\begin{subequations} \label{eq:ude}
    \begin{align}
        \dot{\vec{x}} &= f_p(\vec{x}, \vec{u}(t)) + N\!N_{\theta}(\vec{x}),\\
        \vec{y}_\theta(t) &= h_q(\vec{x}_{p,\theta}(t; \vec{x}_0), t).
    \end{align}
\end{subequations}
Here we use bold variables to denote vectors. $f_p$ are the known (or assumed) dynamics of the system with parameters $p$, $\vec{x}_{p,\theta}(t; \vec{x}_0)$ denotes the solution of those dynamics with initial conditions $\vec{x}_0$ at parameter values $p$ and $\theta$, $\vec{u}(t)$ is an external stimulus, $h_q(\vec{x}_{p,\theta}(t; \vec{x}_0), t)$ is the observation function parameterized by $q$ which can explicitly depend on time (for instance, due to noise processes or the influence of the external stimulus on the observation), and $N\!N_\theta(\vec{x})$ is the universal approximator network trained to learn the unknown dynamics of the system based on its parameters $\theta$. For notational simplicity we drop the parameter dependence of the observation $\vec{y}_\theta(t)$ on parameters $p$ and $q$ and initial conditions $\vec{x}_0$, but keep the network parameters $\theta$, since we want to optimize those while keeping the rest fixed. We focus on systems of ordinary differential equations in this work, but the approach extends to stochastic and delayed differential equation systems, as has been done in previous work using UDEs \cite{rackauckas2021}. For an observed dataset $\{\hat{\vec{y}}_i\}_{i=1, \dots, N}$, training the UDE $N\!N_\theta$ uses a cost function $C(\theta)$ that minimizes the error at every observed time point $t_i$ such that
\begin{equation} \label{eq:udeloss}
    \argmin_\theta C(\theta) = \argmin_\theta \frac{1}{N}\sum_{i=1}^N \left(\vec{y}_{\theta}(t_i)-\vec{\hat{y}}_i\right)^2,
\end{equation}
where $\vec{y}_\theta(t_i)$ is the observed solution of the ODE with neural network weights $\theta$ at the measurement times $t_i$. In all equations that rely on observational data, we denote any observed data for $\vec{x}$ as $\hat{\vec{x}}$. We chose $C(\theta)$ to be the mean squared error in this work, but other cost functions are appropriate for different scenarios, depending on how measurement noise affects the measured signal. We then use the prediction-error method to train UDEs on this loss function for chaotic systems.

\subsubsection{PEM formulation}

The prediction-error method (PEM) is a standard optimization technique for fitting parameters of dynamical systems to observational data \cite{ljung2002, larsson2009}. Similarly to Eq.~\eqref{eq:ude}, we consider a general process $\vec{x}(t)$ that gives rise to some observable state $\vec{y}(t)$ with some measurement noise $\vec{e}(t)$:
\begin{subequations}
    \begin{align}
        \dot{\vec{x}} &= f_p(\vec{x}, \vec{u}(t)),\\
        y_{p,i}(t) &= m_{q,i}(x_{p,i}(t)) + e_{i}(t).
    \end{align}
\end{subequations}
In this statement, $\vec{x}$ is the state of a time-varying system with parameters $p$ that we wish to fit. The observed state $\vec{y}$ is defined by a measurement function $m$ of the process $\vec{x}(t)$, parameterized by $q$, and a measurement error $\vec{e}(t)$. Here we assume that there is a one-to-one mapping of $\vec{x}$ to $\vec{y}$ (i.e., there are no terms in $\vec{y}$ that depend on multiple elements of $\vec{x}$). For a more general discussion of the conditions under which PEM can be applied, see Refs.~\cite{ljung2002, larsson2009}. For example, $m_{q,i}$ is not limited to being a linear function of $x_i$, but can also perform a convolution or incorporate delays \cite{larsson2009}. In all of our examples, a few components of $m_{q,i}$ are zero, indicating that these components, $y_{p,i}$, are not observed. The estimation error $\epsilon$ is weighted by a function $K(t, \beta)$ and added at each time step to the original dynamics $\vec{x}(t)$ to create an adjusted estimate $\tilde{\vec{x}}(t)$ and observations $\tilde{\vec{y}}(t)$ such that
\begin{subequations} \label{eq:pem}
    \begin{align}
        \frac{\mathrm{d}\tilde{\vec{x}}}{\mathrm{d}t} &= f_p(\tilde{\vec{x}}, \vec{u}(t)) + K(t, \beta)\epsilon_p(t),\\
        \tilde{y}_{p,i}(t) &= m_{q,i}(\tilde{x}_{p,i}(t)),\\
        \epsilon_p(t) &= \hat{\vec{y}}(t) - \tilde{\vec{y}}_p(t),
    \end{align}
\end{subequations}
where $\hat{\vec{y}}(t)$ is the observed data interpolated to whatever step is needed by the solver. The error gain factor $K(t, \beta)$ can take on different forms. In this work, we set $K(t, \beta)=K$ as a constant, the simplest form of what previous work has called the parameterized observer \cite{larsson2009}. This observer form is closely related to synchronization-based stabilization of chaotic trajectories \cite{pecora1990}; here it is used inside the UDE training loop rather than as a standalone state-estimation procedure. If there is strong reason to suspect that the quality of the observations varies over time (e.g., much higher noise at certain time intervals), then a time-varying $K(t, \beta)$---for instance an extension of Eq.~\eqref{eq:pem} to resemble an extended Kalman filter where $K(t,\beta)$ is the Kalman gain---might give better performance when fitting. Previous work has nonetheless shown that the parameterized observer outperforms the extended Kalman filter in practice, as tuning the hyperparameter over time is difficult \cite{larsson2009}. For notational simplicity, we will use $K$ in the remaining sections, without loss of generality: an extended Kalman filter can be implemented if there is a data-driven reason to require it.

\subsubsection{Combining PEM-UDE into a single method}

We combine UDE and PEM into a single system by introducing the universal approximator $N\!N_\theta$ to learn the unknown dynamics while simultaneously using the prediction-error feedback to stabilize the training trajectory:
\begin{subequations} \label{eq:pemude}
    \begin{align}
        \frac{\mathrm{d}\tilde{\vec{x}}}{\mathrm{d}t} &= f_p(\tilde{\vec{x}}, \vec{u}(t)) + K\epsilon_\theta(t)+ N\!N_{\theta}(\tilde{\vec{x}}),\\
        \tilde{y}_{\theta,i}(t) &= m_{q,i}(\tilde{x}_{\theta,i}(t)),\\
        \epsilon_\theta(t) &= \hat{\vec{y}}(t) - \tilde{\vec{y}}_\theta(t).
    \end{align}
\end{subequations}
The hyperparameter $K$ controls the loss landscape over which optimization is performed, which we will show is needed to fit UDEs to chaotic systems. In this work, we consider only systems where we have both an initial functional form and observational data. Estimation of unobserved latent states for which we have no initial idea of the functional form is a separate problem discussed elsewhere \cite{chen2022}.

\subsubsection{Symbolic recovery of terms from the PEM-trained UDE}

After fitting the dynamics of the system with the combined PEM-UDE approach, we extract a symbolic form of the dynamics from the universal approximator $N\!N_\theta$. The symbolic form usually generalizes to future contexts more robustly than a black-box neural network \cite{rackauckas2021, cranmer2020}, and allows for direct analysis of the learned functional form. The terms learned by the UDEs can be recovered either by sequential thresholding least squares (STLSQ), which learns features from a library via ridge regression \cite{brunton2016}, or by a genetic algorithm approach to symbolic regression \cite{cranmer2023}. Each has tradeoffs, and the appropriate choice depends on the structure of the data and known dynamics. The genetic algorithm is the better choice when there are complex constraints on the functional form of the system \cite{cranmer2023}, while STLSQ works better on systems that need many data points to achieve an accurate fit. We provide both approaches in the code accompanying this method, and discuss our selection for each of the sections below. Since these approaches are not novel to this work, we leave further discussion to the Methods. In this pipeline, symbolic regression is not used to claim that a neural network is intrinsically interpretable; it is used as a sparsifying projection of the trained residual field, followed by forward simulation of the recovered equations.

\subsection{Analytical properties of the PEM-UDE objective}
\label{sec:theory}
Before the numerical demonstrations, it is useful to make explicit what the prediction-error term does to the optimization problem. Let
\begin{equation}
    F_\theta(\vec{x},t)=f_p(\vec{x},\vec{u}(t))+N\!N_\theta(\vec{x})
\end{equation}
denote the modeled vector field, let $\vec{x}_\theta(t)$ solve the standard UDE, and let $\tilde{\vec{x}}_{\theta,K}(t)$ solve the PEM-UDE with the same initial condition. For a linear observation map $m(\vec{x})=M\vec{x}$, the PEM term may be written as $G(\hat{\vec{y}}-M\tilde{\vec{x}}_{\theta,K})$, with $G=K M^{\!\top}$ for scalar gain and direct component-wise observations. This is the matrix form of Eq.~\eqref{eq:pemude}; for a nonlinear observation map the same local argument holds with $M$ replaced by the observation Jacobian. We compare the ordinary and PEM-assisted trajectory losses in continuous time,
\begin{align}
    C_0(\theta) &= \frac{1}{T}\int_0^T \big\| M\vec{x}_\theta(t)-\hat{\vec{y}}(t)\big\|_2^2\,\mathrm{d}t, \label{eq:costUDE}\\
    C_K(\theta) &= \frac{1}{T}\int_0^T \big\| M\tilde{\vec{x}}_{\theta,K}(t)-\hat{\vec{y}}(t)\big\|_2^2\,\mathrm{d}t, \label{eq:costPEM}
\end{align}
of which Eq.~\eqref{eq:udeloss} is the sampled version.

\paragraph*{Why the minimum is not moved.}
First consider the idealized case that isolates the optimization geometry: the data are noise-free and generated by a member of the model class. Thus there exists $\theta^\star$ such that
\begin{equation}
    \hat{\vec{y}}(t)=M\vec{x}^\star(t),
    \qquad
    \dot{\vec{x}}^\star=F_{\theta^\star}(\vec{x}^\star,t),
\end{equation}
with the same initial condition and unique solutions. At $\theta^\star$, the prediction error is identically zero,
\begin{equation}
    \epsilon_{\theta^\star}(t)=\hat{\vec{y}}(t)-M\tilde{\vec{x}}_{\theta^\star,K}(t)=\vec{0}.
\end{equation}
The PEM feedback therefore switches itself off exactly at the correct dynamics:
\begin{equation}
    \frac{\mathrm{d}\tilde{\vec{x}}_{\theta^\star,K}}{\mathrm{d}t}
    =
    F_{\theta^\star}(\tilde{\vec{x}}_{\theta^\star,K},t).
\end{equation}
By uniqueness, $\tilde{\vec{x}}_{\theta^\star,K}(t)=\vec{x}^\star(t)$, so $C_K(\theta^\star)=0$. Since the loss is a squared norm, $C_K(\theta)\geq 0$ for all $\theta$, and $\theta^\star$ is a global minimizer of the PEM-UDE objective for any fixed gain $K$.

The same argument also shows that PEM leaves the zero-loss set of the observed-state objective unchanged:
\begin{equation}
    \{\theta:C_K(\theta)=0\}=\{\theta:C_0(\theta)=0\}. \label{eq:zeroset}
\end{equation}
If $C_K(\theta)=0$, then the residual vanishes at all times, so the feedback term vanishes along the whole PEM trajectory and that trajectory also solves the unforced UDE. Conversely, any zero-loss standard UDE trajectory has zero residual and is unaffected by the PEM term. The feedback therefore reshapes the landscape away from the data-consistent manifold, but it does not move that manifold. Under structural identifiability and sufficient observability, this common zero set reduces to the true dynamics; under partial observability or model misspecification, different vector fields can still be observationally equivalent, so the statement should be read as preservation of the data-consistent minimum rather than a guarantee of unique global identifiability.

\paragraph*{How PEM makes a chaotic objective trainable.}
Minimum preservation alone does not explain why the optimizer can find the minimum. The gain matters because it changes the stability of the error dynamics. Let $\vec{\xi}(t)=\tilde{\vec{x}}_{\theta,K}(t)-\vec{x}^\star(t)$, define the model-error field
\begin{equation}
    \vec{\delta}_\theta(\vec{x},t)=F_\theta(\vec{x},t)-F_{\theta^\star}(\vec{x},t),
\end{equation}
and let $J^\star(t)=D_{\vec{x}}F_{\theta^\star}(\vec{x}^\star(t),t)$. Linearizing the PEM-UDE around the true trajectory gives
\begin{equation}
    \dot{\vec{\xi}}
    =
    \underbrace{\left[J^\star(t)-GM\right]}_{A_K(t)}\vec{\xi}
    +\vec{\delta}_\theta(\vec{x}^\star(t),t)
    +\mathcal{O}(\|\vec{\xi}\|^2).
    \label{eq:errdyn}
\end{equation}
For full observation with scalar gain, $GM=K I$, so the PEM term shifts each exponent of the linearized error dynamics by $-K$. The unstable direction with exponent $\lambda_{\max}$ becomes contracting when
\begin{equation}
    K>\lambda_{\max}. \label{eq:Kcond}
\end{equation}
This gives a simple physical interpretation of the loss-landscape smoothing: the observer damps the exponentially growing separation between nearby trajectories before it contaminates the loss and its gradients. In the R\"{o}ssler example, the Lyapunov time is $\lambda_{\max}^{-1}\approx 10$, or $\lambda_{\max}\approx 0.1$, and the empirically useful gains $K\approx0.18$--$0.25$ lie just above this threshold, consistent with Fig.~\ref{fig:s3}. For partial observation, the same damping occurs in the observed unstable subspace; unobserved unstable directions must be coupled to the observations for the feedback to stabilize them.

Let $\Phi_K(t,s)$ be the state-transition matrix generated by $A_K(t)$. If the observed error dynamics are uniformly contracting, so that $\|\Phi_K(t,s)\|\leq C\exp[-\gamma_K(t-s)]$ for $t\geq s$, variation of constants gives
\begin{equation}
    \vec{\xi}(t)
    =
    \int_0^t \Phi_K(t,s)\vec{\delta}_\theta(\vec{x}^\star(s),s)\,\mathrm{d}s
    +\mathcal{O}(\|\vec{\xi}\|^2),
    \qquad
    \|\vec{\xi}(t)\|
    \lesssim
    \frac{C}{\gamma_K}
    \sup_{0\leq s\leq t}\|\vec{\delta}_\theta(\vec{x}^\star(s),s)\|,
    \label{eq:residual}
\end{equation}
where $\gamma_K=K-\lambda_{\max}$ for full observation with scalar gain.
The PEM loss near the minimum is therefore controlled by the autonomous vector-field error rather than by exponentially diverging trajectories:
\begin{equation}
    C_K(\theta)
    \lesssim
    \frac{C^2\|M\|^2}{(K-\lambda_{\max})^2}
    \sup_{0\leq t\leq T}\|\vec{\delta}_\theta(\vec{x}^\star(t),t)\|^2.
    \label{eq:costbound}
\end{equation}
Thus $K$ converts a chaotic trajectory-matching problem into a stable, observer-regularized comparison of vector fields, while Eq.~\eqref{eq:zeroset} keeps the exact minimum fixed. The trade-off is visible in the same scaling: increasing $K$ suppresses sensitive dependence and smooths the landscape, but it also compresses differences between nearby vector fields by the factor $(K-\lambda_{\max})^{-2}$, producing the plateau seen at large gains.

\paragraph*{Noisy observations and gain selection.}
With measurement noise, $\hat{\vec{y}}(t)=M\vec{x}^\star(t)+\vec{e}(t)$, the linearized error dynamics become
\begin{equation}
    \dot{\vec{\xi}}
    =
    A_K(t)\vec{\xi}
    +\vec{\delta}_\theta(\vec{x}^\star(t),t)
    +G\vec{e}(t).
\end{equation}
The observer now damps chaotic divergence but also injects measurement noise into the simulated state. Let $\mathcal{L}_K$ denote convolution with the stable transition operator $\Phi_K$. To leading order, the observed residual is
\begin{equation}
    M\tilde{\vec{x}}_{\theta,K}-\hat{\vec{y}}
    \approx
    M\mathcal{L}_K\vec{\delta}_\theta
    +\left(M\mathcal{L}_K G-I\right)\vec{e},
\end{equation}
and, for zero-mean noise independent of the model error,
\begin{equation}
    \mathbb{E}\,C_K(\theta)
    \approx
    \frac{1}{T}\int_0^T
    \left\|M\mathcal{L}_K\vec{\delta}_\theta\right\|_2^2\,\mathrm{d}t
    +
    \mathbb{E}\left\|\left(M\mathcal{L}_K G-I\right)\vec{e}\right\|_2^2 .
    \label{eq:noisyCK}
\end{equation}
The first term is the useful contrast between different autonomous vector fields; the second term is a $K$-dependent noise floor. Small gains leave the chaotic instability active. Very large gains make the observer track the measured time series so strongly that model errors are hidden by feedback and the informative contrast in Eq.~\eqref{eq:noisyCK} becomes shallow. This motivates a useful interval,
\begin{equation}
    \lambda_{\max}<K\lesssim K_{\mathrm{noise}},
\end{equation}
where the lower bound stabilizes the trajectory and the upper scale is set by the point at which the model-error term becomes comparable to the noise floor. The precise upper scale depends on the observation operator and the spectrum of the measurement noise, so it is not universal. In practice, when a Lyapunov scale can be estimated from data, we choose $K$ just above that scale and check the resulting loss landscape; for the neural mass model we instead use a continuation schedule and anneal $K$ from $0.3$ to $0.1$ before symbolic extraction, reducing the residual influence of the observer on the learned autonomous dynamics.

\paragraph*{What the neural network term learns.}
The network is not intended to learn the whole vector field. In the missing-physics formulation it learns only
\begin{equation}
    g(\vec{x},t)=F_{\theta^\star}(\vec{x},t)-f_p(\vec{x},\vec{u}(t)),
\end{equation}
the part not already represented by the known physics. This smaller target is central to why a small network can work and why a large network is not automatically better. For noisy data, the fitted residual field has the usual bias-variance structure,
\begin{equation}
    \mathbb{E}\!\left[\|N\!N_{\hat{\theta}}(\vec{x})-g(\vec{x})\|_2^2\right]
    =
    \underbrace{\|\overline{N\!N}(\vec{x})-g(\vec{x})\|_2^2}_{\mathrm{bias}^2}
    +
    \underbrace{\mathbb{E}\!\left[\|N\!N_{\hat{\theta}}(\vec{x})-\overline{N\!N}(\vec{x})\|_2^2\right]}_{\mathrm{variance}},
    \label{eq:biasvar}
\end{equation}
where $\overline{N\!N}$ is the noise-averaged learned network. Increasing the network size can reduce the bias, but it increases the variance term, which is the part that follows a particular noise realization.

This capacity dependence is the standard bias--variance trade-off \cite{hastie2009}. A local linearization illustrates it explicitly. If $\Theta\in\mathbb{R}^{N\times P}$ is the resulting feature matrix, $P$ is the number of active degrees of freedom, and $\sigma_{\mathrm{eff}}^2$ is the effective noise variance in the regression target, then with ridge/weight-decay regularization
\begin{equation}
    \frac{1}{N}\sum_{i=1}^N
    \mathrm{Var}\!\left[N\!N_{\hat{\theta}}(\vec{x}_i)\right]
    \approx
    \frac{\sigma_{\mathrm{eff}}^2}{N}
    \mathrm{tr}\!\left[
    \Theta(\Theta^{\!\top}\Theta+\lambda I)^{-1}\Theta^{\!\top}
    \right]
    =
    \frac{d_{\mathrm{eff}}\sigma_{\mathrm{eff}}^2}{N},
    \qquad
    d_{\mathrm{eff}}\leq P .
    \label{eq:variance}
\end{equation}
The noise-fitting term therefore grows with the effective number of parameters and falls with the number of informative samples. The optimal model size is finite: it is the smallest network for which the bias has become small compared with the variance cost. Because $f_p$ already contains the dominant physical structure, the residual $g$ is much lower-complexity than the full dynamics, so this optimal size can be small. This is why we use compact $N_1\times10\times10\times N_2$ networks rather than large black-box neural ODEs.

These arguments mitigate, but do not eliminate, the risk that the network absorbs dynamics attributed to the known physics. In our experiments $f_p$ and its assumed parameters are held fixed, so the network target is the residual needed to match the observed trajectory. The data nevertheless constrain the total field mainly on the sampled states: a flexible network can compensate for misspecified known terms or adopt different off-trajectory extensions. Keeping $P$ small and using AdamW weight decay restricts those alternatives through an objective of the form
\begin{equation}
    C_{K,\lambda}(\theta)=C_K(\theta)+\lambda R(\theta).
    \label{eq:regobj}
\end{equation}
Finally, the symbolic regression step imposes an additional sparsity constraint and the learned symbolic equations are validated after the PEM term is reduced or removed. These checks reduce, but cannot rule out, residual non-identifiability away from the sampled attractor. They also help explain why STLSQ succeeds on PEM-UDE outputs but fails on the raw noisy data in Table~\ref{tab:ppstlsq}: integration through the trained UDE produces a smooth generative trajectory, whereas numerical differentiation of noisy data amplifies high-frequency noise by a factor proportional to $\omega^2$ in power.

\subsection{PEM-UDE outperforms standard UDE in fitting chaotic systems}

\begin{figure*}
    \centering
    \includegraphics[width=\textwidth]{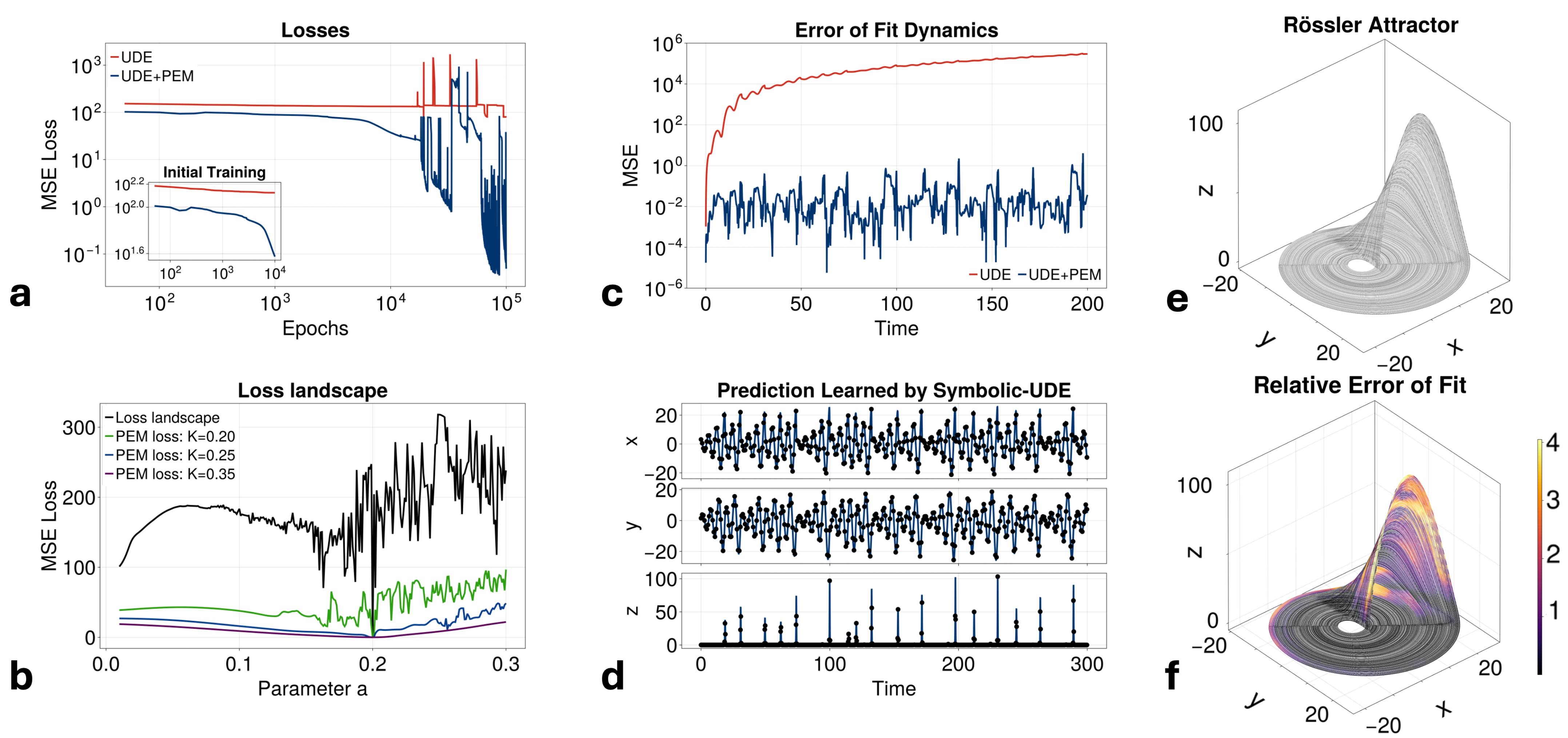}
    \caption{The combined PEM-UDE approach can learn the dynamics of chaotic systems that traditional UDEs cannot. \textbf{(a)} Comparison of training error for the UDE system (fails to capture dynamics accurately) and the PEM-trained UDE (succeeds). The difference in accuracy is already large 10\% into the full training session (inset), at which point the unassisted UDE is no longer progressing. \textbf{(b)} The application of PEM smooths an otherwise difficult parameter loss landscape, with the hyperparameter $K$ tuning the steepness. \textbf{(c)} Mean-squared error of the fit dynamics, UDE versus PEM-trained UDE. \textbf{(d)} Using STLSQ to replace the PEM-trained UDE produces an expression that captures the original dynamics (first 200 time points) and continues to describe them accurately well into the future (last 100 time points). \textbf{(e)} The surface of the R\"{o}ssler attractor shows chaotic filling of phase space, with a basin near the $x$-$y$ plane and occasional bursts along the $z$-axis. \textbf{(f)} Relative error of the PEM-trained UDE at each point of the attractor surface.}
    \label{fig:2}
\end{figure*}

We begin by showing how the PEM-UDE approach fits the R\"{o}ssler system (Fig.~\ref{fig:2}). This system was chosen because it is a simple set of ODEs with only a single nonlinear term, yet it produces chaotic dynamics for the right parameter set \cite{rossler1976}. In this example, we assume that we know the functional form of $x$ and $z$ but have no information about the functional form of $y$, and so must learn the dynamics from the time series observations. With only this challenge, a traditional UDE fails to fit well (Fig.~\ref{fig:2}a): the UDE-learned dynamics typically converge to the mean of the time series, a local minimum in the objective function. The PEM-UDE approach, by contrast, converges successfully after training, with a clear difference in performance even after the first 10\% of the total training time, at which point the standard UDE has already neared its local minimum.

The reason for this performance increase lies in the way that the PEM approach alters the landscape over which the UDE is trained (Fig.~\ref{fig:2}b). The unmodified landscape is extremely rough and contains several local minima, with only a very sharp decrease in loss at the true global minimum ($a=0.2$ in this case). When PEM is added to the dynamics of the system during training of the UDE, the landscape becomes much more amenable to standard optimization techniques. The choice of $K$ also tunes the steepness of the landscape directly. This should be done in conjunction with considering how the PEM will smooth out the landscapes of other parameters in states where it is not applied. In the noise-free realizable case, PEM preserves the data-consistent zero-loss set, but it can flatten the landscape in directions associated with unobserved states or misspecified parameters. As higher levels of filtering are introduced with larger $K$, finding the true functional form of the entire system depends more on the accuracy of the estimates of the unfit parameters (Fig.~\ref{fig:s2}). With the PEM-UDE combination approach, then, the best choice of $K$ is one that produces a smooth enough landscape for optimization but is still steep enough that the global minimum is clearly distinguishable from the surrounding area, so that noisier estimates of unmeasured parameters are acceptable.

The PEM-UDE approach works because the prediction error correction removes the sensitive dependence on parameters from the system being fit, neutralizing the defining feature of chaotic systems. As shown in Fig.~\ref{fig:s3}a, when $K$ is small there is an $\mathcal{O}(1)$ divergence with respect to any deviation in the initial parameter choice (illustrated as a deviation of $10^{-9}$ in Fig.~\ref{fig:s3}a), but when $K$ is sufficiently large ($K\approx0.18$ in this example) this sensitive dependence is eliminated and the error between the two trajectories shrinks. For sufficiently large $K$, the residual error depends on the size of the deviation in the parameter, and $K$ can be tuned based on the estimate of this deviation (Fig.~\ref{fig:s3}b). For noise-free observations generated within the model class, the feedback vanishes on the data-consistent trajectory and preserves the zero-loss set [Eq.~\eqref{eq:zeroset}]. With noise, partial observability, or misspecification, PEM instead trades stability against a shallower and potentially biased objective [Eq.~\eqref{eq:noisyCK}]. The $L_2$-norm trajectory fitting approach, which is unstable on the original chaotic system, is thereby stabilized over the range of gains examined here.

The dynamics learned by the PEM-UDE fit the system well, while the initial estimate of the UDE alone does not offer an accurate prediction even within the Lyapunov time of the system (Fig.~\ref{fig:2}c). For the parameters selected here, the Lyapunov time of the system is $\lambda^{-1} \approx 10$, with most nearby areas of parameter space having similar characteristic times ($\lambda^{-1} < 25$; see Methods for further details). The mean error of the UDE alone continues to grow throughout the simulation time, whereas the mean error at each timestep of the PEM-UDE fit dynamics is two to three orders of magnitude smaller than the dynamics themselves.

To generalize these dynamics beyond the original observations, we extracted a symbolic form of the equations using an STLSQ fit to the outputs of the UDE taken at each time step of the dynamics of the PEM-UDE system. The functional form of the missing state is directly recovered by STLSQ ($\dot{y} = p_1x + p_2y$), with the initial fits of parameters close to the true values ($p_1 = 1.01$, $p_2=0.17$, while the true values are $p_1=1$, $p_2=0.2$). An additional optimization step that tunes the parameters of this learned symbolic equation after STLSQ converges to the true values (accurate to three decimal places), which then generalize the dynamics forward in time beyond the original simulations (Fig.~\ref{fig:2}d). The pipeline thus provides an analytical form of the dynamics that generalizes forward in time and can also be examined for insights into the physical system, recovering the original attractor of the system with an error typically under ~4\% (Fig.~\ref{fig:2}e-f). This benchmark therefore evaluates hidden-state recovery, parameter error, out-of-window simulation, and attractor geometry, rather than only short-horizon trajectory matching.

\subsection{PEM-UDE can learn chaotic dynamics under severe observational noise}

\begin{figure*}
    \centering
    \includegraphics[width=\textwidth]{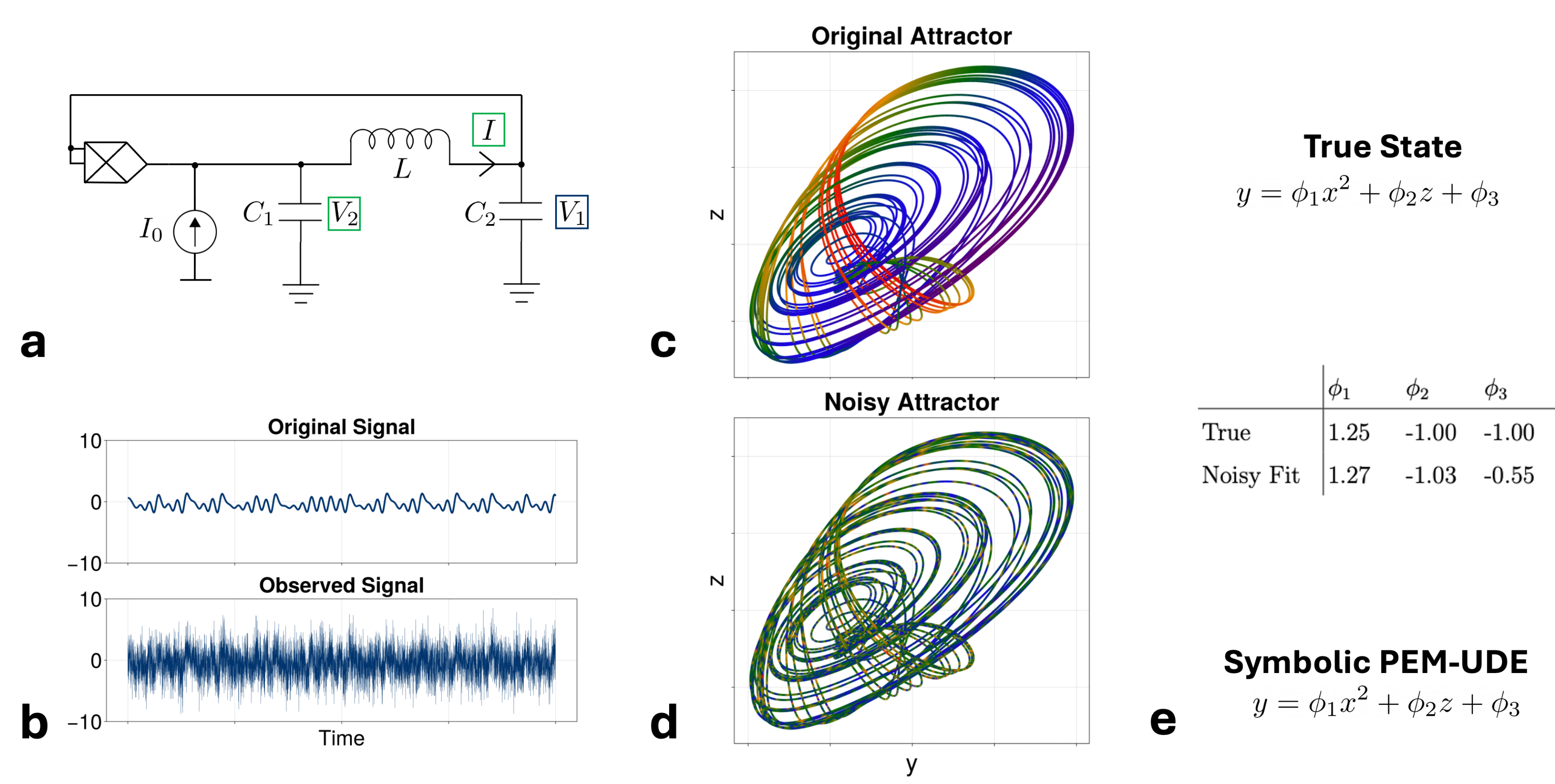}
    \caption{The PEM-UDE approach is capable of recovering the correct form of a chaotic circuit even in the presence of significant observational noise. \textbf{(a)} The analog Petrzela--Polak circuit that produces the dynamics examined in this section. The observed states are the dimensionless forms of $V_1$, $V_2$, and $I$, where $V_2$ and $I$ (green boxes) are observed with only a small amount of noise, and $V_1$ (blue box) is observed with large observational noise. \textbf{(b)} In this example system, we assume a faulty observation of one state ($V_1$), where the true dynamics (top) are masked by observational noise of $5\times$ the true signal's magnitude (bottom). \textbf{(c)} Original attractor surface, with the true form of the observed dimension $x$ in color and the magnitude of error from observational noise shown. \textbf{(d)} Attractor surface fit by the PEM-UDE method with noise in the third dimension (the scattered colors indicate that the noise has significantly corrupted the dynamics of the state $x$). \textbf{(e)} The true form of the state to be recovered is shown at the top, with the form recovered by STLSQ applied to the PEM-UDE dynamics shown at the bottom. The functional form, although not the exact parameters, is recovered from the PEM-UDE fit (table compares parameters fit by STLSQ to original parameters).}
    \label{fig:3}
\end{figure*}

We next present a second example of the PEM-UDE method being fit to a chaotic system, with two important differences from the R\"{o}ssler system. First, this example is a physically realized electrical circuit, with the circuit diagram shown in Fig.~\ref{fig:3}a and the parameters for the physical circuit (as described in Ref.~\cite{sprott2022}) given in Table~\ref{tab:ppparams}. The move from the idealized chaotic dynamics of the previous system to a real-world circuit also has practical relevance, since chaos in electrical circuits has applications of its own \cite{kuznetsov2023}. Second, we add a noisy observation to one of the circuit components ($V_1$, denoted by the blue box in Fig.~\ref{fig:3}a). The true equations for this system are given in Eq.~\eqref{eq:ppcirc}, with the dimensionless form used in this example given by Eq.~\eqref{eq:ppbase}. In this section, we assume there is an unknown faulty observation of the first state $x$, such that the noise around the observation is $5\times$ the magnitude of the true signal (Fig.~\ref{fig:3}b). This creates a difficulty for the PEM-UDE fit: the true dynamics (Fig.~\ref{fig:3}c) are now masked by enough noise to distort the shape of the attractor being fit (Fig.~\ref{fig:3}d).

\begin{table}
\caption{Direct symbolic regression with STLSQ applied to noisy Petrzela--Polak circuit data fails to recover the correct dynamics under the same data and noise conditions where the PEM-UDE approach succeeds.}
\label{tab:ppstlsq}
\begin{ruledtabular}
\begin{tabular}{ll}
$\lambda$ & Fit expression \\ \hline
0.1  & $\phi_1 + \phi_2x+\phi_3y+\phi_4z+\phi_5xy+\phi_6y^2+\phi_7xz+\phi_8z^2$ \\
0.2  & $\phi_1 + \phi_2x+\phi_3y+\phi_4z+\phi_5xy+\phi_6xz+\phi_7yz+\phi_8z^2$ \\
0.3  & $\phi_1+\phi_2y+\phi_3z+\phi_4z^2$ \\
0.4  & $\phi_1+\phi_2y+\phi_3z+\phi_4y^2+\phi_5yz$\\
0.5  & $\phi_1+\phi_2y+\phi_3z$
\end{tabular}
\end{ruledtabular}
\end{table}

We assume that this noise is unknown at the time of fitting and proceed with fitting the PEM-UDE to the corrupted data using the same approach as above (fitting the functional form shown in Eq.~\eqref{eq:pplearn}). Even with the substantial observational noise in one of the states, STLSQ applied to the PEM-UDE fit dynamics recovers the exact functional form of the dynamics (Fig.~\ref{fig:3}e). The approach does not provide the original parameters, although the form is correct. The problem has effectively been shifted from one of unknown dynamics to one of simple parameter optimization, to which other approaches (e.g., simulation-based inference \cite{goncalves2020}) can be applied. This indicates that even with very noisy observational data, the PEM-UDE approach can recover useful information about the true dynamics of the system.

This example also illustrates why fitting the symbolic form of a trained UDE outperforms an STLSQ fit of the original dataset. As a comparison to the dynamics shown in Fig.~\ref{fig:3}e, we fit the same noisy dataset directly using the SINDy approach \cite{brunton2016, juliusmartensen2021}: we compute a dataset of smoothed derivative estimates and perform the STLSQ symbolic regression method directly on the derivative estimates. As shown in Table~\ref{tab:ppstlsq}, this method cannot recover the true symbolic form across a range of the ADMM hyperparameter $\lambda$, with more sparse representations incorrectly dropping the noisy state $x$ from the expression entirely. Recovering the symbolic form of the trained UDE is more accurate because the UDE provides a noise-free (or greatly noise-reduced) version of the dynamics that can be sampled with arbitrary density, sidestepping the sensitivity that symbolic regression methods show to noisy data. This comparison is not meant to exhaust all possible smoothing or weak-form variants of SINDy; rather, it isolates a failure mode that is common in derivative-based symbolic discovery, where differentiating noisy observations amplifies high-frequency error while integrating the learned UDE produces a smooth generative trajectory for subsequent sparse regression.

\subsection{Learning novel expressions for sparse networks that violate analytical derivation assumptions}

\begin{figure*}
    \centering
    \includegraphics[width=0.80\textwidth]{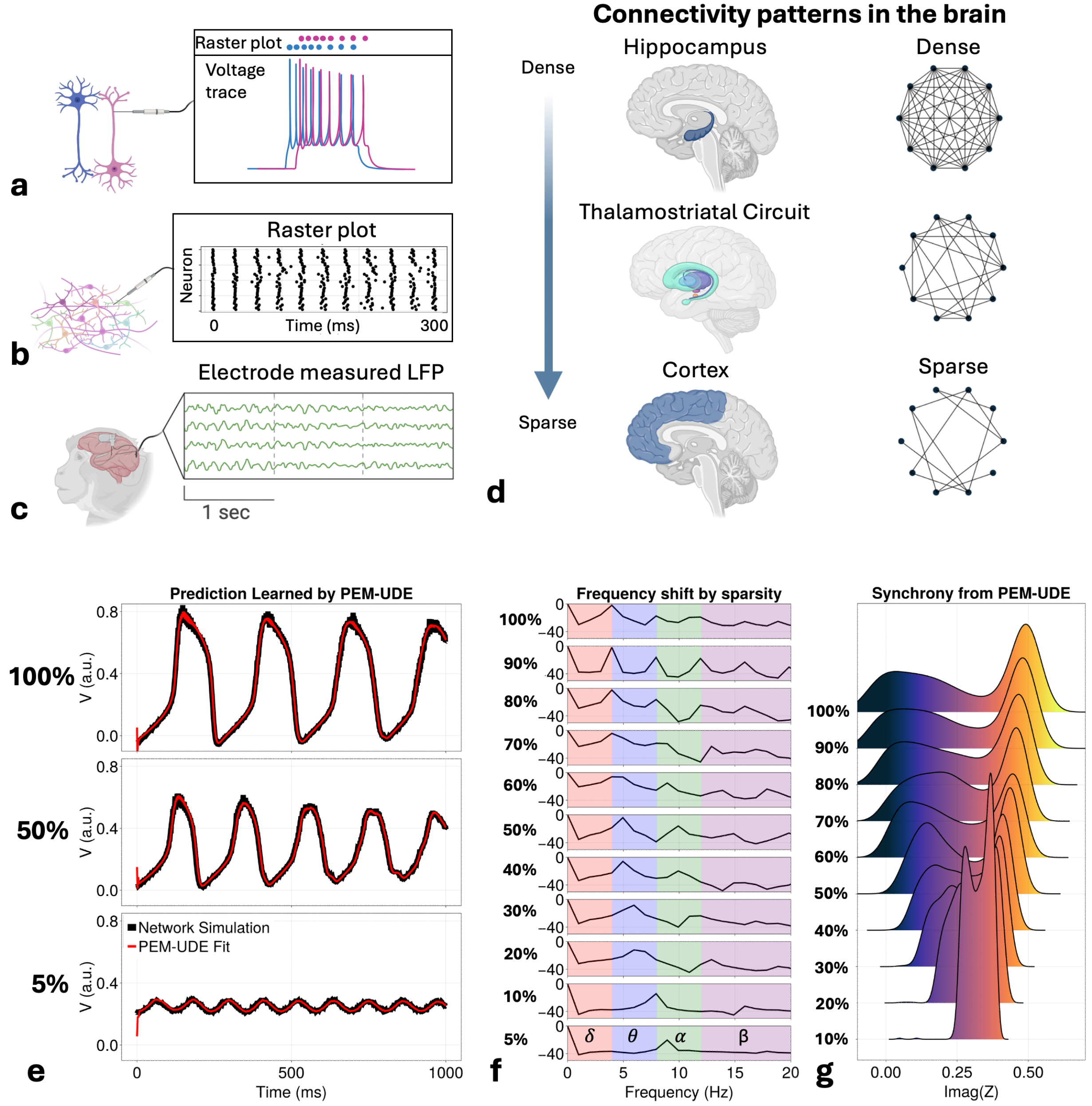}
    \caption{Learning neural mass models that account for regional heterogeneity in neural connection sparsity helps explain the origin of different frequency bands within the brain. Single neurons produce spike trains \textbf{(a)}, which are encoded as a raster plot of individual dots for each neuron/spike. In a population, this view makes group activity easy to read, and synchronized regimes show clearly visible bursts \textbf{(b)}. In behavioral experiments, electrode arrays can be placed to record population-level activity from multiple brain regions \textbf{(c)}. The degree of local connectivity within brain regions varies across the brain \textbf{(d)}, with deeper regions exhibiting greater connectivity and cortical regions having sparser within-region connections. To account for this sparsity, we train a next-generation neural mass model to learn terms that capture the emergent population dynamics as connections become more sparse \textbf{(e)}. The PEM-UDE learned dynamics exhibit qualitatively similar frequency trends to those seen in real brain regions, with a shift from theta to alpha as the dominant frequency as network connections become increasingly sparse \textbf{(f)}. Higher sparsity also limits the synchronization that is possible within the network \textbf{(g)}. Figure created with BioRender.com.}
    \label{fig:4}
\end{figure*}

Having shown that this approach can learn the dynamics of chaotic physical systems, we next turn to deriving expressions for the mean-field activity of neuron populations. In this section the learned equations are reduced-order closure equations for simulated microscopic networks, not equations fit directly to the experimental recordings considered later. The data needed to study the human brain spans multiple scales: individual neuronal activity gives rise to synchronized actions that define interacting functional circuits (Fig.~\ref{fig:4}a--c). Traditional neural mass models represent the average activity of neuron populations by modeling particular phenomena (e.g., excitatory/inhibitory balance in the Wilson--Cowan model \cite{wilson1972}, ion-gradient dynamics in excitatory populations in the Larter--Breakspear model \cite{larter1999, breakspear2003}), but lack a direct tie to many of the biophysical parameters (e.g., neurotransmitters) needed to understand inter-neuron dynamics. Next-generation neural mass models (NGNMMs) address this by providing an analytical derivation of the mean-field dynamics that retains the biophysical details of individual neurons and also captures the emergent properties at the population level \cite{montbrio2015, chen2022}. The derivation and its key assumptions are discussed further in the Methods.

The analytical form of mean-field dynamics offered by NGNMMs has several advantages over traditional neural mass models. First, there is a direct connection between the single-neuron parameters and the dynamics of the neural population, a feature only available in certain populations of traditional neural masses tuned to be sensitive to these parameters \cite{larter1999, breakspear2003}. Second, NGNMMs capture \textit{emergent properties} of neural populations that traditional neural masses do not. For example, $\beta$-rebound, where suppressed activity in the $\beta$ band increases in power after being suppressed (a phenomenon often observed in motor cortex studies), is observed as a feature of coupled NGNMMs that are built only to oscillate in the $\beta$ band, with no functional constraints that force this activity to occur \cite{coombes2019}.

There are also limitations to the NGNMM approach, mostly involving the assumptions needed for the Ott--Antonsen ansatz \cite{ott2008a}, a key component of the NGNMM derivation. Most consequential from a biophysical standpoint is the assumption of all-to-all connectivity of the neuron population. This is approximately correct in deep regions (e.g., hippocampus, entorhinal cortex) that have significant within-region collaterals \cite{chen2022}. Immediate neighbors of these regions tend to have fewer collaterals, however, as in the thalamostriatal connection patterns \cite{ding2010}. The assumption of full connectivity becomes obviously false in the cortex \cite{pathak2024}, which typically has very sparse connectivity (on the order of 1--5\% of all possible connections; Fig.~\ref{fig:4}d). Accounting for sparsity in NGNMMs is therefore a problem that must be solved for them to be useful in new contexts.

There is disagreement on the correct analytical approach to adjust the derivation of NGNMMs for sparsity \cite{divolo2018, goldobin2021, divolo2022, clusella2024}, so we sidestep the issue of an analytical derivation and learn the form of the sparse networks from simulated microscopic-network data with the PEM-UDE method. Following the analytical approach for Izhikevich neurons \cite{chen2022}, we begin with a large population of neurons described by Eq.~\eqref{eq:izhnrn} with synaptic dynamics given by Eq.~\eqref{eq:izhnrnsyn}. In the all-to-all case, the analytical NGNMM describing the activity of this population is given by Eq.~\eqref{eq:ccbase}. As we begin to prune connections in the graph to introduce sparsity, this form no longer accurately represents the population activity. We therefore learn correction terms for the mean firing rate $r$ and voltage $v$ given a reduced probability of connectivity $p_c$, where $p_c \in [0.05, 1]$. Although there are additional states for the recovery current $w$ and the synaptic dynamics $s$, we only learn from the firing rate and voltage states and train only on their data, since these are the data typically available from experiments. Restricting training to these two macroscopic states does not make the later experimental recordings the training set; it instead mirrors the observability constraints that population recordings impose.

The PEM-UDE approach captures the mean-field dynamics of the network under different degrees of sparsity with good accuracy (Fig.~\ref{fig:4}e). This accuracy does not include the initial conditions; this is discussed further in the limitations portion of the Discussion. The method does capture a shift in the dominant frequency of oscillations present in the network as a function of sparsity (Fig.~\ref{fig:4}f). Densely connected regions typically have a primary frequency peak in the $\theta$-band, while the most sparse networks exhibit a primary peak in the $\alpha$-band. This emerges as a natural feature of the learned dynamics and is qualitatively consistent with experimental data. Hippocampal rhythms are typically in $\theta$ frequencies and emerge from densely connected areas of the brain \cite{lubenov2009}, while $\alpha$ frequencies typically emerge from sparsely connected cortical regions before propagating through the brain \cite{halgren2019}. The PEM-UDE learned dynamics therefore extend the strength of NGNMMs in capturing emergent properties (e.g., synchrony) at the mean-field scale by accurately capturing the ergodic properties of the system. The learned dynamics also generalize beyond the training data: when trained only on $p_c = 0.3$--$0.9$, the PEM-UDE still recovers dynamics down to 5\% connectivity (Fig.~\ref{fig:s4}).

\begin{figure*}
    \centering
    \includegraphics[width=\textwidth]{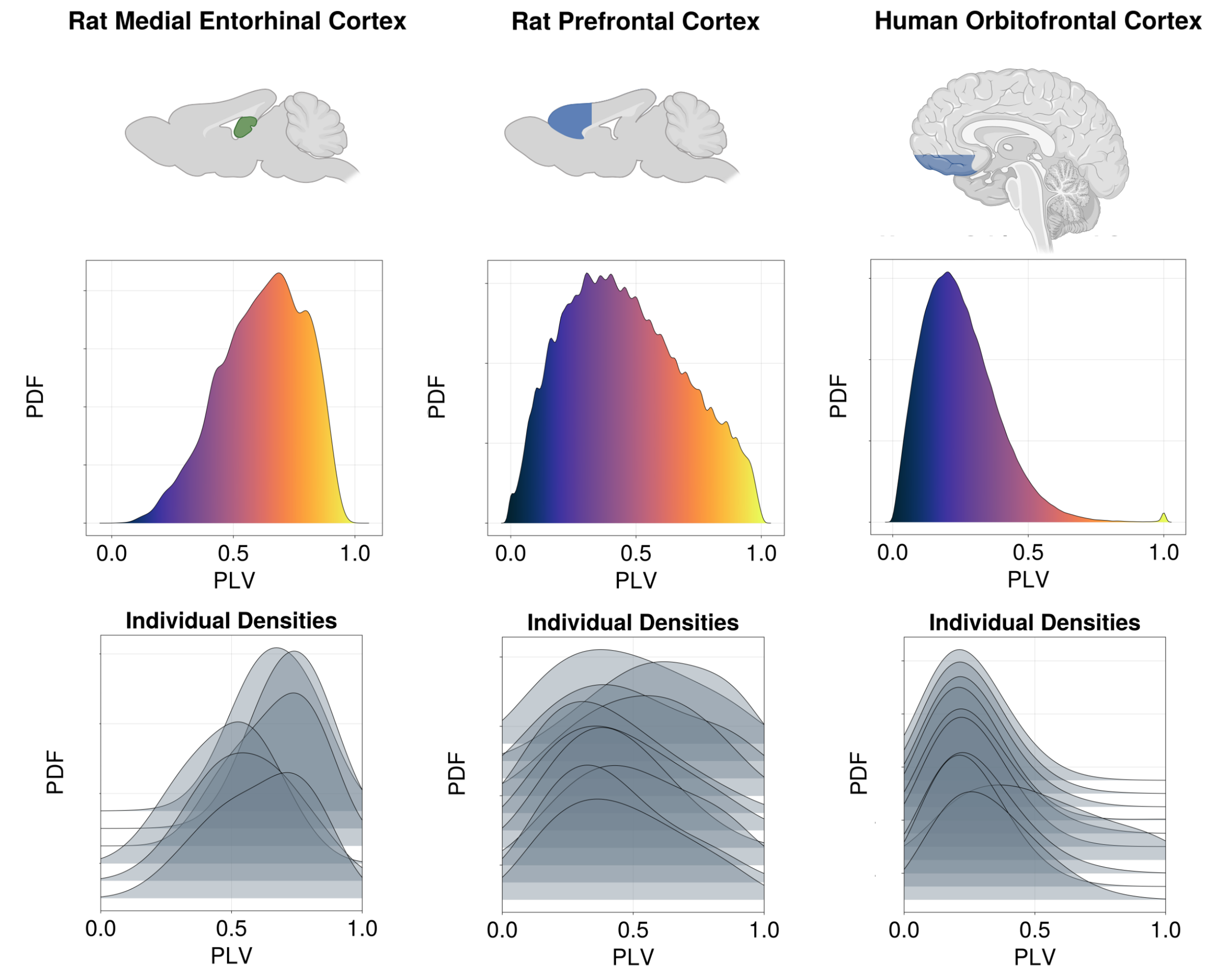}
    \caption{Intracranial EEG shows synchrony variation across regions consistent with the symbolic PEM-UDE-derived neural mass model. As shown in Fig.~\ref{fig:4}g, the typical synchrony in regions with greater connectivity is higher than in regions with lower connectivity. A similar ordering appears in biological brains: we examined three datasets. In the region with the greatest connectivity (rat medial entorhinal cortex; left column), there is a greater degree of synchrony (measured by phase-locking value; PLV) over the duration of the experiment than in the prefrontal cortex (middle column). The difference is larger still compared to the human orbitofrontal cortex at rest (right column), where overall synchronization tends to be very low during spontaneous activity. As all cumulative distributions (upper row) are based on synchrony values from many individuals, synchrony distributions from each subject are also shown (bottom row) to illustrate the variability across individuals. Figure created with BioRender.com.}
    \label{fig:5}
\end{figure*}

As with the previous two systems, we extend the PEM-UDE learned dynamics described in Eq.~\eqref{eq:cclearn} by fitting a symbolic form of the dynamics to the UDE network with STLSQ, giving an analytical form of the solution (Eq.~\eqref{eq:ccsymb}, with fit parameters in Table~\ref{tab:ccfit}). Izhikevich and related quadratic-integrate-and-fire NGNMMs can undergo period-doubling and chaotic collective regimes under other parameter choices and coupling structures \cite{nicola2013,schmidt2020,nandi2024}. The parameter set used here, however, lies in a stable population-oscillation regime. The neural example therefore tests the same training framework in a partially observed oscillatory system and is distinct from the two chaotic benchmarks. Since NGNMMs are derived based on the assumption of a population of coupled oscillators, there is an exact expression for the degree of synchrony within the population given by the Kuramoto order parameter $Z$:
\begin{subequations}
\begin{align}
    Z &= \frac{1-W^*}{1+W^*},\\
    W &= \pi r + iv.
\end{align}
\end{subequations}
For the PEM-UDE learned network dynamics, we observe an emergent prediction that more densely connected regions tend to have higher degrees of synchrony as a baseline state, while more sparsely connected regions tend to exhibit lower degrees of synchrony when not acted upon by an external drive (Fig.~\ref{fig:4}g). This is an unexpected prediction from the learned reduced-order equations for networks of different sparsity. To test it, we examined three datasets of intracranial electrode recordings from regions of different degrees of sparsity (Fig.~\ref{fig:5}). Because these datasets differ in species, anatomical region, electrode configuration, task context, and behavioral state, this comparison cannot isolate local connection probability as the causal variable. It is instead an external consistency check: regions whose known anatomy implies different typical local connectivity show the same ordering of resting synchrony predicted by the simulated reduced-order model. These datasets included recordings from the rat medial entorhinal cortex (MEC; $N=6$), a deep region near the hippocampus with high intraregional connectivity \cite{hernandez-perez2020, hernandez-perez2020a}; the rat prefrontal cortex ($N=10$), which has significantly sparser connectivity \cite{peyrache2009, peyrache2018}; and human orbitofrontal cortex ($N=10$), which exhibits a similarly high degree of sparsity \cite{saez2018, saez2018a}. Consistent with the learned reduced-order equations, the deeper, more densely connected structures exhibit higher synchrony than either of the cortical datasets, and the pattern holds across individuals, although with significant variability between subjects. These results support the plausibility of the model's qualitative prediction outside the simulation set, while leaving causal validation of the learned terms to future experiments with directly measured connectivity.

\section{Discussion}

Here we introduce a method to fit universal differential equations in chaotic systems and use it to learn reduced-order equations for simulated neural population activity. The approach uses the prediction error method to train the UDE system, and a symbolic form of the learned dynamics is then extracted via symbolic regression \cite{brunton2016}. We show that the method can learn the dynamics of a classical chaotic system, the R\"{o}ssler system \cite{rossler1976}. A standard UDE alone cannot learn this system effectively, but the combined PEM-UDE approach captures the dynamics accurately, suggesting that the method has identified the true functional form of the dynamics rather than projecting within conventional statistical boundaries. Prior work in UDEs has focused on stiff systems \cite{kim2021}, fitting UDEs in the presence of stochastic noise \cite{oleary2022}, and extracting explainable functions from UDE networks \cite{rackauckas2021}. Separately, prior work in parameter optimization using the PEM has shown its utility in fitting parameters of dynamical systems \cite{ljung2002, larsson2009}, and that it can even outperform neural network fits of dynamics \cite{sun2021, pillonetto2025}. Our methodological contribution is the use of prediction-error feedback inside UDE training for chaotic trajectories, followed by recovery and autonomous validation of an explicit residual field; observers, UDEs, and symbolic regression themselves are established methods. The approach succeeds where standard UDE methods fail because it smooths the parameter landscape that the UDE must optimize. Under the noise-free realizability and uniqueness assumptions stated in Sec.~\ref{sec:theory}, this deformation preserves the data-consistent zero-loss set; under noise, misspecification, or incomplete observability, the analysis instead identifies the conditions and trade-offs that limit that statement. We also extend the approach to a second chaotic system from an electrical circuit \cite{petrzela2019, sprott2022}, and show that even in the presence of significant observational noise, the PEM-UDE method can recover the true functional form of an unknown state. More accurate estimation of the parameters of these recovered dynamics could be obtained by taking the functional form learned from the PEM-UDE dynamics and applying a more advanced optimization technique that fully exploits the ability to simulate the system in different regimes (e.g., simulation-based inference \cite{goncalves2020}).

The analytical results of Sec.~\ref{sec:theory} explain these empirical observations while delimiting their scope. Prediction-error feedback leaves the zero-loss set unchanged for noise-free realizable data because the feedback vanishes on the data-consistent manifold [Eq.~\eqref{eq:zeroset}]. For full observation with scalar gain, it shifts the linearized error exponents by $-K$ [Eq.~\eqref{eq:errdyn}], so $K>\lambda_{\max}$ is a local sufficient condition for contraction. Partial observation additionally requires coupling of unobserved unstable directions to measured states. With noise or misspecification, the same analysis shows why the gain must remain moderate: strong feedback can mask autonomous model error. The bias--variance argument of Eqs.~\eqref{eq:biasvar}--\eqref{eq:variance} motivates the compact networks used here, but does not by itself guarantee identifiability away from the sampled attractor.

The cost of the observer term is small relative to the UDE solve-and-differentiate loop: it adds interpolation of the measured state and a vector addition to the right-hand side at solver timesteps, while the dominant computation remains ODE integration, sensitivity calculation, and network optimization. We therefore view PEM-UDE as a stabilization of standard UDE training rather than a competing state-estimation method. Other approaches developed for chaotic identification, including multiple shooting, weak-form objectives, shadowing-based losses, and filtering-assisted symbolic regression, are natural complementary baselines and can often be combined with the same observer-regularized UDE. Recent SINDy-type work has likewise improved robustness to noisy data by combining linear multistep approximations, ensemble adaptive libraries, and inner-product sparse regression \cite{jiang2026}. Our comparisons establish performance relative to an otherwise identical UDE and to direct STLSQ under the tested conditions; they do not establish state-of-the-art performance across all chaotic-identification methods. A systematic cross-method benchmark over higher-dimensional, stiff, and multiscale systems remains future work.

Having shown that the PEM-UDE approach can learn the dynamics of chaotic systems, we then considered how to learn reduced-order governing equations in systems of neurons under biologically plausible constraints. These constraints are of two kinds: \textit{experimental} constraints on the data available, which require using only data sources that can be feasibly acquired \textit{in vivo} (firing rate, membrane voltage); and \textit{neurobiological} constraints, where the sparsity of interneuron connections must be accurately modeled. Experimental constraints are worth keeping in mind because they make the method more immediately applicable, but methodological advances can often relax these over time. Biological constraints, and connection density in particular, are more important to model accurately, as they shape the kinds of information processing possible within the brain \cite{balcioglu2023, fruengel2025}. We use a missing-physics approach, where the initial form of the population activity is given by an NGNMM \cite{montbrio2015, chen2022}. NGNMMs capture emergent properties of neuron populations and retain the microscale parameters that are typically manipulated \textit{in vitro} \cite{coombes2019, sheheitli2024}, but they rely on the assumption of all-to-all connectivity, which is non-physiological for most brain regions \cite{montbrio2015}. There have been varied attempts to sidestep this limitation and include corrections for sparsity by reweighting the degree of synaptic conductance or the distribution of synaptic currents, with varying degrees of success \cite{divolo2018, goldobin2021, divolo2022, goldobin2019, clusella2024}. Each of these approaches comes with new assumptions and its own limitations. Here, we instead learn a reduced-order closure for an NGNMM of a simulated population of Izhikevich neurons directly from the observations of the network itself. The closure equations learned this way predict emergent properties that are observed directly in biology, such as the shift from $\theta$ to $\alpha$ as the dominant frequency band as networks become more sparse, a feature seen in the brain moving from densely connected regions (e.g., hippocampus) to sparsely connected regions (e.g., prefrontal cortex) \cite{lubenov2009, halgren2019}. The equations also predict that more densely connected regions exhibit greater intraregional synchrony at rest than sparsely connected networks. To test this hypothesis, we computed the intraregional synchrony from three datasets of intracranial electrode recordings (rat MEC, rat prefrontal cortex, and human orbitofrontal cortex) \cite{hernandez-perez2020, peyrache2009, saez2018}. The ordering seen in these datasets is consistent with the simulated model's prediction, but because connectivity was not measured directly it should be interpreted as a biological consistency check rather than causal validation.

The neural-population result should be interpreted at the level of a reduced-order closure. The microscopic simulations use Izhikevich neurons and parameter values inherited from the analytical NGNMM construction of Chen and Campbell \cite{chen2022}, chosen to represent hippocampal-like dynamics under the all-to-all limit. The learned terms then correct that mean-field system as the connectivity probability is varied. This is different from common statistical fits of neural recordings, such as generalized linear or point-process models of spike trains \cite{paninski2004, truccolo2005}: those models are often simpler and more identifiable for encoding or decoding experiments, while PEM-UDE aims to preserve a biophysical population-level state space that can retain microscopic parameters and emergent synchrony. The price of that added structure is that the learned sparsity corrections should not be read as unique biological mechanisms until they are tested against recordings with matched structural connectivity or prospectively varied microscopic parameters. A decisive biological test would pair recordings with measured structural connectivity or controlled perturbations within a common preparation.

The method also has limitations that we will address in future work. Very noisy systems can be difficult to learn, especially since chaotic systems can be hard to average over trials. Care must also be taken: nonphysical solutions can shadow physical ones \cite{chandramoorthy2021}. The inclusion of information during training via the PEM, together with the extraction of a symbolic form of the dynamics, helps to mitigate these difficulties. In models that include known physics, PEM can lead to close-but-inaccurate minima if the parameters are not exactly known. This limits the method's applicability to scenarios where the parameters included in the known physical model are sufficiently accurate. In principle, this can be overcome by fitting to a range of simulations across these parameters, but the training time will increase substantially. The method is also best suited to situations with a known structure of data (i.e., some driving inputs are known). Neural experiments often provide this level of accuracy, but there is also substantial interest in resting-state data \cite{nozari2024}. Adapting the approach to fit stationary time series without known perturbations is therefore another extension we see as important for neuroscience applications.

For the neural example specifically, the frequency shift with sparsity may depend on the chosen neuron parameters, synaptic scaling, graph ensemble, finite-size effects, delays, heterogeneity, and external drive. We therefore interpret the theta-to-alpha transition as a prediction of the simulated Izhikevich network family studied here, not as a universal law for all neural populations. The comparison with rat MEC, rat prefrontal cortex, and human orbitofrontal cortex is likewise limited by species, region, electrode, behavioral, and clinical differences; future validation would ideally use recordings paired with measured structural connectivity within a common preparation.

Although the work presented here shows the utility of this approach in neuroscience, there are several areas for further development. Future work can incorporate loss objectives that include optimization of initial conditions, which can help generalization to real-world systems \cite{sun2021, ji2021}. Future work could also focus on learning ``black-box'' representations of the dynamics, where symbolic regression methods are not applied to learn the dynamics. This can be useful when complexity penalties (i.e., enforced sparsity within the learned expression) bias away from the correct dynamics. For example, when many small inputs lead to complex interactions based on their temporal pattern, the high-dimensional representation present in the neural network may be more desirable. The temporal sequences of visual cortex projections to the prefrontal cortex are one example of this kind of structured high-dimensional input space \cite{pathak2024}. In such a case, rather than fitting the dynamics symbolically, a second optimization run of the UDE without PEM after the initial combined run could lead to more generalizable results (since this forces the contribution of PEM to 0 rather than letting it minimize naturally over the optimization problem). Including a method for directly estimating stochastic processes within the dynamics, as has been done in some standard UDE approaches \cite{oleary2022}, would add another layer of generalization to the approach. This is especially relevant for finite or partially observed neural populations: the quenched input distribution in the NGNMM appears through $\bar{\eta}$ and $\Delta$, but unobserved neurons can also generate stochastic finite-size fluctuations that are not represented by the deterministic closure used here. A stochastic PEM-UDE would let those fluctuations be learned as process noise rather than forcing the deterministic residual terms to absorb them. Extending the PEM-UDE approach to systems of partial differential equations would also allow for direct fitting of neural field models under different degrees of connection density, which is of increasing interest as neuroscience tries to understand the role of traveling waves in the brain \cite{roberts2019, breakspear2017} and how the relationship between densely and sparsely connected regions shapes the flow of cognitive processes \cite{balcioglu2023}.

\section{Methods}

\subsection{Scientific machine learning in the Julia ecosystem}

The Julia ecosystem has a solid infrastructure for scientific machine learning \cite{rackauckas2021} that we use to build the PEM-UDE method described here. Since setting up and training UDEs in Julia has been discussed elsewhere (Refs.~\cite{rackauckas2021, kim2021} are good examples), we refer interested readers to the accompanying code for implementation details and give only a few brief technical notes here on the choices of solvers, network structure, fitting procedure, and symbolic regression.

In this work, we default to non-stiff solvers (typically Vern7), mostly for speed. As long as the UDE network is large enough to allow for a decent approximation of the unknown dynamics and the system itself is not very stiff, a non-stiff adaptive solver should give sufficiently accurate fits. If the system is known or strongly suspected to be stiff, then switching to a stiff solver (e.g., Rodas4) is recommended and is easy in the provided code, since the Julia ecosystem has a wide range of solvers available. This switch may become necessary either because chaotic systems are often stiff in certain regions, or because even when the system itself is not stiff (as the neural systems discussed here can be in the correct parameter space), the inclusion of a universal approximator will often lead to a stiff problem, at least until the initial parameter weights are adjusted. We recommend increasing the size of the universal approximator network before defaulting to a stiff solver, as this can have a large impact on performance.

We choose the radial basis function (RBF; notation given in Ref.~\cite{craddock1996}) as the activation function for the universal approximator network, given by
\begin{equation}
    \phi(x) = \exp(-x^2).
\end{equation}
The RBF is chosen for both theoretical and practical reasons. There is a substantial literature showing that the RBF is useful in the context of universal approximation \cite{craddock1996, park1991, zhao2019}. Universal-approximation results motivate this function class but do not make a finite network size arbitrary. For noisy missing-physics problems, larger networks are not automatically preferable: once the residual field can be represented, additional capacity increases the number of directions in which the fit can follow a particular noise realization. We therefore use a compact architecture and treat capacity as a regularization choice. For the networks shown in this work, we use a depth of four layers, where for an input dimension $N_1$ and output dimension $N_2$, each network has size $N_1 \times 10 \times 10 \times N_2$.

As the PEM portion of training requires interpolating the original training data to the solver time step, we use DataInterpolations.jl to do simple linear interpolation between observations \cite{bhagavan2024}. The additional numerical work introduced by PEM is therefore the interpolation call and the feedback addition in the right-hand side; it does not add a separate filtering pass or a second optimization problem. Although there are many optimizers provided to train the UDEs, we have defaulted to using the AdamW method (decoupled weight decay and gradient updates) \cite{loshchilov2019} in all of the results presented here.

The symbolic recovery of terms uses sequential thresholded least squares (STLSQ) solved using the alternating direction method of multipliers (ADMM) \cite{boyd2010}, as implemented in DataDrivenDiffEq.jl \cite{juliusmartensen2021}. Following Ref.~\cite{brunton2016}, $\Theta$ is the feature library with weights $\xi$ such that $y=\Theta\xi$, with sparsity enforced via LASSO \cite{tibshirani1996}:
\begin{equation}
\argmin_\xi \tfrac{1}{2}\|\Theta\xi-y\|_2 + \lambda\|\xi\|_1.
\end{equation}
As STLSQ is faster than genetic algorithm approaches for large datasets ($N \gtrsim 10{,}000$) \cite{cranmer2023}, we default to this method. In principle, the trained UDE can generate subsamples of arbitrary density, so either approach can be used.

\subsection{Background on the R\"{o}ssler system}

The R\"{o}ssler system is a set of three nonlinear differential equations originally introduced as a simplified model for studying chaotic dynamics. It contains only one nonlinear term but still produces complex chaotic behavior. The dynamics are described by the 3D ODE system:
\begin{subequations} \label{eq:rossbase}
    \begin{align}
        \dot{x} &= -y-z,\\
        \dot{y} &= x + ay,\\
        \dot{z} &= b + z(x-c).
    \end{align}
\end{subequations}
For specific parameter values ($a=0.2$, $b=0.22$, $c=14$ in the case presented here), the system produces the funnel-shaped R\"{o}ssler attractor, which exhibits sensitivity to initial conditions, fractal structure, and period-doubling chaotic activity. We chose this as an initial test of the PEM-UDE combination because, despite its simplicity, the R\"{o}ssler system has chaotic properties including positive Lyapunov exponents, topological entropy, and various bifurcation sequences that have been used to study transitions between ordered and chaotic regimes in dynamical systems. The R\"{o}ssler system has also been extended to analytically describe certain physical problems (e.g., mechanical vibrations, atmospheric variability).

In the results presented here, we use a universal approximator $U$ to learn the dynamics of the entire second state, assisted by the observational data through the PEM technique. The system simulated during learning is therefore given by
\begin{subequations} \label{eq:rosslearn}
    \begin{align}
        \dot{x} &= -y-z,\\
        \dot{y} &= U(\theta, x, y, z) + K\cdot \left[\hat{y}(t) - y(t)\right],\\
        \dot{z} &= b + z(x-c).
    \end{align}
\end{subequations}
For the results presented in the main text, we use $K=0.25$. This choice is informed by the steepness of the parameter landscapes discussed in the Results; other reasonable values of this hyperparameter will not affect the accuracy of the learned dynamics, provided an appropriate solver is chosen. The results here are based on 100,000 training epochs with the AdamW optimizer \cite{loshchilov2019}. The loss function used to train the UDE network is the mean-squared error between the predicted states at each time step and the true time series (with a small amount of noise added for robustness during training):
\begin{equation} \label{eq:genloss}
    \ell(\theta) = \|\hat{\vec{x}} - \vec{x} \|^2.
\end{equation}
For the parameters selected in our analyses, we characterized the degree of chaos via the Lyapunov time of the system as $\lambda^{-1} \approx 10$. Most parameter combinations in the surrounding space have similar values ($>80\%$ have $\lambda^{-1} \in [0, 25]$; see supplemental code for computation of the empirical CDF of the maximum Lyapunov exponents). Lyapunov exponents were computed using standard functions offered in ChaosTools.jl \cite{datseris2018, datseris2022}.

\subsection{Background on the Petrzela--Polak circuit}

Petrzela and Polak \cite{petrzela2019} introduced a family of analog electrical circuits exhibiting chaotic dynamics under three constraints: a canonical form, structurally stable attractors, and novel dynamics. We work with a particular realization whose chaotic dynamics can be physically realized in a simple circuit, with the attractor observable on an oscilloscope (demonstrated in Ref.~\cite{sprott2022}, from which we take the parameters and non-dimensional form discussed here). For the circuit shown in Fig.~\ref{fig:3}a, the equations describing the three observed states are
\begin{subequations} \label{eq:ppcirc}
    \begin{align}
        \dot{V}_1 &=\frac{I}{C_1},\\
        \dot{V}_2 &= \frac{I_m - I - I_0}{C_2},\\
        \dot{I} &= \frac{V_2-V_1-IR_L}{L}.
    \end{align}
\end{subequations}
For convenience in numerical simulations, we work with the non-dimensional form of the system (i.e., one in which physical constants are replaced with the minimal number of parameters to represent the same dynamics):
\begin{subequations} \label{eq:ppbase}
    \begin{align}
        \dot{x} &= z,\\
        \dot{y} &= ax^2 - z - c,\\
        \dot{z} &= y - x - bz.
    \end{align}
\end{subequations}
All parameters for both the dimensional and dimensionless systems are taken from Ref.~\cite{sprott2022} and are provided in Table~\ref{tab:ppparams}.

\begin{table}
\caption{Physical and non-dimensional parameters of the Petrzela--Polak circuit.}
\label{tab:ppparams}
\begin{ruledtabular}
\begin{tabular}{ll}
\textbf{Parameter} & \textbf{Value} \\ \hline
$R_L$           & 0.5\,k$\Omega$ \\
$R_M$           & 0.08\,k$\Omega$ \\
$C_1$           & 1.0\,$\mu$F \\
$C_2$           & 1.0\,$\mu$F \\
$L$             & 1.0\,H \\
$I_0$           & 1.0\,mA \\ \hline
\textbf{Dimensionless parameter} & \\ \hline
$a$           & 1.25 \\
$b$           & 0.5 \\
$c$           & 1.0 \\
\end{tabular}
\end{ruledtabular}
\end{table}

The form of the equations learned by the PEM-UDE method is given by:
\begin{subequations} \label{eq:pplearn}
    \begin{align}
        \dot{x} &= z,\\
        \dot{y} &= U(\theta, x,y,z) + K\cdot\left[\hat{y}(t) - y(t)\right],\\
        \dot{z} &= y - x - bz.
    \end{align}
\end{subequations}
The loss function used to train this PEM-UDE combination is the same as Eq.~\eqref{eq:genloss}, to test the robustness of the system to a poorly observed state. If it is known \textit{a priori} that the observations of a state are corrupted to that degree, it could be worth modifying the loss in Eq.~\eqref{eq:genloss} to weight the error in this state less during training, or to drop the observations altogether. As shown in the previous section, the PEM-UDE training method can recover dynamics even when some states are completely hidden, provided the initial guess of their dynamics is sufficiently accurate.

\subsection{Spiking neural networks and next-generation neural mass models}

Next-generation neural mass models rely on the Ott--Antonsen (OA) ansatz to derive the mean-field activity of a population of neurons \cite{ott2008a, montbrio2015}. The OA ansatz was introduced as a way to bypass the need to set up an entire Fokker--Planck equation for the mean-field activity of a population of coupled oscillators, provided certain assumptions are met. The OA ansatz \cite{ott2008a} assumes that the population of coupled oscillators exists in the thermodynamic limit (i.e., $N \rightarrow \infty$) and that the probability distribution of the observable (typically voltage) in phase space converges to a Lorentzian. The derivation can approximate other distributions to a reasonable degree of accuracy \cite{divolo2018}. The OA ansatz approach also assumes that the noise input into each neuron receives a heterogeneous, quenched noise input $\eta$ drawn from a general distribution. The currents cannot be stochastic; otherwise, the Lorentzian ansatz is violated. The OA ansatz further assumes that the conditional probabilities of the macroscopic variables are separable. This assumption and its limitations have been examined in detail \cite{goldobin2019}, and a second-order circular cumulant approximation has been derived for neural mass models \cite{divolo2022}. The OA ansatz also makes the assumption we address in this paper: the network being approximated has an all-to-all connectivity structure. As discussed above, this is a non-physiological assumption for most regions of the brain, and removing it is needed to develop more accurate neural mass models.

We do not require the OA ansatz for the work here beyond the initial assumption of the functional form for population activity that will be used to train the UDE, so we omit most of the derivation and give only the main points. The equations shown here are adapted from earlier work that gives the entire derivation \cite{chen2022}. The parameter set is inherited from that work and is intended to represent hippocampal-like Izhikevich population dynamics in the all-to-all limit; we do not tune these microscopic parameters to the experimental datasets. Following that work, we simulate a network of Izhikevich neurons that take the form
\begin{subequations} \label{eq:izhnrn}
\begin{align}
    \dot{V}_j &= V_j(V_j - \alpha) -w_j +\eta_j +I_\text{ext} + I_{\text{syn,}j}, \\
    \dot{w}_j &= a(bV_j-w_j),
\end{align}
\end{subequations}
where $V_j$ and $w_j$ are the voltage and recovery current, respectively, of neuron $j \in [1,N]$ for a network of $N$ neurons. The voltage activity in these neurons depends on the internal neuronal dynamics, the distributed noise source $\eta_j$, any external driving current $I_\text{ext}$, and the synaptic current from other interneuronal connections $I_{\text{syn},j}$. The synaptic activity is given by
\begin{subequations} \label{eq:izhnrnsyn}
\begin{align}
    I_{\text{syn,}j} &= g_\text{syn}s(e_r-V_j), \\
    \dot{s}_j &= -\frac{s_j}{\tau_s} + \frac{s_\text{jump}}{N} \sum_{k=1}^N{\sum_{t_k<t}\delta(t-t_k)}.
\end{align}
\end{subequations}
Each neuron therefore receives an impulse of $s_\text{jump}/N$ at every time a presynaptic neuron $k$ fires. If the network is structured to fulfill the assumptions discussed above, the mean-field activity of the entire population can be derived using the OA ansatz and takes the form
\begin{subequations} \label{eq:ccbase}
\begin{align}
    \dot{r} &= \frac{\Delta}{\pi} + 2rv - (\alpha + g_\text{syn}s)r,\\
    \dot{v} &= v(v-\alpha) - w + \bar{\eta} + I_\text{ext} + g_\text{syn}s(e_r-v)-(\pi r)^2,\\
    \dot{w} &= a(bv-w) + w_\text{jump}r,\\
    \dot{s} &= -\frac{s}{\tau_s} + s_\text{jump}r,
\end{align}
\end{subequations}
where $r$ is the population firing rate, $v$ the mean voltage, $w$ the mean recovery current, and $s$ the mean synaptic activity. As discussed in earlier work, this is an accurate representation of the mean-field activity for fully connected networks \cite{chen2022}. Although this model class admits chaotic regimes for other parameter choices, the parameters in Table~\ref{tab:ccorig} produce the stable population oscillations studied here.

For all simulations presented here, we simulate a network of 1000 Izhikevich neurons with varying degrees of network sparsity, where the probability of two neurons forming a synapse is given by $p_c \in\{0.05, 0.1, 0.2, \ldots, 1.0\}$ as indicated for the sparsities shown in Fig.~\ref{fig:4}. In a standard network of neurons, as the sparsity of connections is varied, the synaptic currents must be rescaled in proportion to the median in-degree of connectivity $K$ to maintain balanced (synchronous) activity. Previous work has shown that rescaling the synaptic conductance with $g \propto 1/\sqrt{K}$, with similar rescaling of background currents, provides accurate scaling for a broad spectrum of network arrangements (in-degrees with Lorentzian, Gaussian, and Erd\H{o}s--R\'enyi distributions) \cite{divolo2018}. Following that work, we rescale synaptic conductance with $p_c$ such that
\begin{equation}
    s_\text{jump} = \frac{s_{\text{jump},0}}{\sqrt{p_c}},
\end{equation}
with $s_{\text{jump},0}=s_\text{jump}\sqrt{1.5}\approx1.5$. As in the previous sections, we set up the UDE to learn the unknown terms for the states we are interested in, but with two differences. We learn terms for both the first and second states in this derivation and so apply PEM to both of these states. We also modify Eq.~\eqref{eq:genloss} to train on only data for the first two states, since these are the only states typically observable in experimental data. The initial parameters for this system are given in Table~\ref{tab:ccorig}, and the formal description of the PEM-UDE system is given by
\begin{subequations} \label{eq:cclearn}
\begin{align}
    \dot{r} &= \frac{\Delta}{\pi} + 2rv - (\alpha + g_\text{syn}s)r \nonumber\\
    &\quad + U_1(\theta, r, v, w, s) + K_1\cdot \left[\hat{r}(t) - r(t)\right],\\
    \dot{v} &= v(v-\alpha) - w + \bar{\eta} + I_\text{ext} + g_\text{syn}s(e_r-v)-(\pi r)^2 \nonumber \\
    &\quad+ U_2(\theta, r, v, w, s) + K_2\cdot\left[\hat{v}(t) - v(t)\right],\\
    \dot{w} &= a(bv-w) + w_\text{jump}r,\\
    \dot{s} &= -\frac{s}{\tau_s} + s_\text{jump}r.
\end{align}
\end{subequations}
After learning the dynamics with the PEM-UDE approach and recovering the symbolic form of the UDE expressions via STLSQ, the new NGNMM with correction factors for sparsity is given by the parameter values in Table~\ref{tab:ccfit} and by the functional form:
\begin{subequations} \label{eq:ccsymb}
\begin{align}
    \dot{r} &= \frac{\Delta}{\pi} + 2rv - (\alpha + g_\text{syn}s)r + f_1(\harpoon{u}),\\
    \dot{v} &= v(v-\alpha) - w + \bar{\eta} + I_\text{ext} + g_\text{syn}s(e_r-v)-(\pi r)^2 \nonumber\\
    &\quad+ f_2(\harpoon{u}),\\
    \dot{w} &= a(bv-w) + w_\text{jump}r,\\
    \dot{s} &= -\frac{s}{\tau_s} + s_\text{jump}r,\\
    f_1(\harpoon{u}) &= \phi_1 s + \phi_2 rs + \phi_3 vs + c_1,\\
    f_2(\harpoon{u}) &= \phi_4 r + \phi_5 rv + \phi_6 rs + \phi_7 s^2 + c_2.
\end{align}
\end{subequations}
This form is fit only to the sparsity at 50\% connectivity (see Fig.~\ref{fig:s5} in the Supplemental Material) and requires additional tuning to different sparsities. The Supplemental Material now gives the complete neural-fit sequence---training data and observables, optimizer and iteration count, gain continuation, symbolic extraction, extrapolation test, and interpretation of the learned terms---together with the limits on mechanistic interpretation.

\begin{table}
\caption{Parameters of the Izhikevich neuron population and NGNMM, from Ref.~\cite{chen2022}.}
\label{tab:ccorig}
\begin{ruledtabular}
\begin{tabular}{lll}
\textbf{Parameter} & \textbf{Value} & \textbf{Interpretation} \\ \hline
$\alpha$                        & 0.6215         & Dimensionless Izhikevich voltage parameter \\
$g_\text{syn}$                  & 1.2308         & Synaptic conductance \\
$I_\text{ext}$                  & 0              & External current \\
$e_r$                           & 1              & Synaptic reversal potential \\
$a$                             & 0.0077         & Izhikevich recovery current parameter \\
$b$                             & $-0.0062$        & Izhikevich recovery current parameter \\
$w_\text{jump}$                 & 0.0189         & Recovery current update per spike \\
$\tau_s$                        & 2.6            & Synaptic time constant \\
$s_\text{jump}$                 & 1.2308         & Synaptic activity update per spike \\
$v_\text{peak}, v_\text{reset}$ & 200            & Cutoffs for $v_\infty$ \\
$\bar{\eta}$                    & 0.12           & Mean of Lorentzian for applied noise \\
$\Delta$                        & 0.02           & Width of Lorentzian for applied noise \\
\end{tabular}
\end{ruledtabular}
\end{table}

\begin{table}
\caption{Parameters of the symbolic PEM-UDE-learned NGNMM at $p_c=0.5$.}
\label{tab:ccfit}
\begin{ruledtabular}
\begin{tabular}{lll}
\textbf{Parameter} & \textbf{Value} & \textbf{Interpretation} \\ \hline
$\phi_1$           & $-0.50$         & Reduction of synaptic contribution to firing rate \\
$\phi_2$           & $-0.12$         & Rescaling of $g_\text{syn}$ \\
$\phi_3$           & $0.88$          & Voltage-dependent synaptic activity \\
$\phi_4$           & $-5.01$         & Scaling of firing rate contribution to voltage \\
$\phi_5$           & $9.80$          & Nonlinear interaction: firing rate $\times$ voltage \\
$\phi_6$           & $-0.05$         & Nonlinear interaction: firing rate $\times$ syn. \\
$\phi_7$           & $-0.27$         & Quadratic synaptic contribution to voltage \\
$c_1$              & $0.12$          & Rescaling of $\Delta$ \\
$c_2$              & $0.06$          & Rescaling of $\bar{\eta}$ \\
\end{tabular}
\end{ruledtabular}
\end{table}

\subsection{Experimental data acquisition and preprocessing}

Acquisition and preprocessing of each of the three experimental datasets are described in detail in their respective publications \cite{hernandez-perez2020, peyrache2009, saez2018}; we describe them only briefly here. The rat MEC dataset \cite{hernandez-perez2020a} was acquired from six male Long--Evans rats, with electrode recordings taken during a long period of roaming in a circular track. The rat prefrontal cortex dataset \cite{peyrache2018} was acquired from four male Long--Evans rats during 10 trials in which they successfully learned a behavioral task, including the rest periods before and after the task recordings. The human orbitofrontal cortex dataset \cite{saez2018a} was acquired during electrocorticography recordings of 10 neurosurgical patients undergoing evaluation for epilepsy. The patients played a decision-making game, and the recordings used for this analysis are the inter-trial rest portions of the ECoG data. For all datasets, the synchrony between electrodes within the region was computed as the phase-locking value (PLV) between the angle of Hilbert-transformed signals ($\phi$) from each electrode \cite{kuramoto1984, mormann2000}. Following prior work \cite{pathak2024}, spike data from the datasets that recorded spike times rather than LFPs were converted to an LFP proxy by windowed spike counts, with the PLV calculation performed on these instead. PLV is given by
\begin{equation}
\text{PLV} = \left| \frac{1}{N}\sum_{t=1}^N e^{i[\phi_i(t) - \phi_j(t)]}\right|
\end{equation}
for all discretized $t \in [1, N]$ and all pairs of electrodes $(i, j)$.

\begin{acknowledgments}
We thank Alex Driussi and Botond Antal for their valuable support with graphics.
The research presented here was funded by the Baszucki Brain Research Fund (L.R.M.-P.).
A.G.C.\ acknowledges the NIGMS MSTP Training Award T32-GM008444.
\end{acknowledgments}

\section*{Data and code availability}

All code and simulated data to reproduce the results and plots in this manuscript, along with an additional abbreviated demonstration for interested readers to rapidly prototype with, can be found in the GitHub repository associated with this publication at \url{https://github.com/Neuroblox/pem-ude}. The intracranial electrode recording datasets can all be accessed in repositories on CRCNS.org, with the specific conditions of access to each one listed on their individual repository pages \cite{hernandez-perez2020a, peyrache2018, saez2018a}.

\section*{Author contributions}

Conceptualization: A.G.C., D.H., C.V.R., L.R.M.-P., H.H.S.\
Methodology: A.G.C., D.H., V.D., C.V.R., H.H.S.\
Software: A.G.C., D.H., V.D.\
Formal analysis: A.G.C., D.H., H.H.S.\
Investigation: A.G.C., D.H.\
Resources: E.K.M., R.H.G., A.E., L.R.M.-P., H.H.S.\
Data curation: A.G.C., D.H.\
Writing---original draft: A.G.C., D.H., H.H.S.\
Writing---review and editing: all authors.\
Visualization: A.G.C., D.H.\
Supervision: C.V.R., L.R.M.-P., H.H.S.\
Funding acquisition: L.R.M.-P., H.H.S.

\section*{Competing interests}

C.V.R., E.K.M., R.H.G., A.E., L.R.M.-P., and H.H.S.\ are co-founders of Neuroblox Inc., a company spun out of Stony Brook University, MIT, and Dartmouth College to develop a commercial-grade software platform for multi-scale computational neuroscience with applications to diagnosis and treatment of brain-based disorders. The other authors declare no competing interests.

\bibliography{references_ant_no_lang_v5}

@article{sejnowski_computational_1988,
  author  = {Sejnowski, Terrence J. and Koch, Christof and Churchland, Patricia S.},
  title   = {Computational Neuroscience},
  journal = {Science},
  volume  = {241},
  number  = {4871},
  pages   = {1299--1306},
  year    = {1988},
  doi     = {10.1126/science.3045969}
}

@book{strogatz_nonlinear_2015,
  author    = {Strogatz, Steven H.},
  title     = {Nonlinear Dynamics and Chaos: With Applications to Physics, Biology, Chemistry, and Engineering},
  edition   = {2nd},
  publisher = {Westview Press},
  year      = {2015},
  isbn      = {978-0813349107}
}

@article{strogatz2000,
  author  = {Strogatz, Steven H.},
  title   = {From {Kuramoto} to {Crawford}: Exploring the Onset of Synchronization in Populations of Coupled Oscillators},
  journal = {Physica D: Nonlinear Phenomena},
  volume  = {143},
  number  = {1--4},
  pages   = {1--20},
  year    = {2000},
  doi     = {10.1016/S0167-2789(00)00094-4}
}

@article{poirazi_illuminating_2020,
	title = {Illuminating dendritic function with computational models},
	volume = {21},
	copyright = {2020 Springer Nature Limited},
	issn = {1471-0048},
	url = {https://www.nature.com/articles/s41583-020-0301-7},
	doi = {10.1038/s41583-020-0301-7},
	
	number = {6},
	urldate = {2025-11-12},
	journal = {Nature Reviews Neuroscience},
	author = {Poirazi, Panayiota and Papoutsi, Athanasia},
	month = jun,
	year = {2020},
	note = {Publisher: Nature Publishing Group},
	pages = {303--321},
}

@article{abbott_realistic_1991,
	title = {Realistic synaptic inputs for model neural networks},
	volume = {2},
	issn = {0954-898X},
	url = {https://doi.org/10.1088/0954-898X_2_3_002},
	doi = {10.1088/0954-898X_2_3_002},
	number = {3},
	urldate = {2025-11-12},
	journal = {Network: Computation in Neural Systems},
	author = {Abbott, L F},
	month = jan,
	year = {1991},
	pages = {245--258},
}

@article{goldobin2021,
  title = {Reduction Methodology for Fluctuation Driven Population Dynamics},
  author = {Goldobin, Denis S. and di Volo, Matteo and Torcini, Alessandro},
  journal = {Phys. Rev. Lett.},
  volume = {127},
  issue = {3},
  pages = {038301},
  numpages = {6},
  year = {2021},
  month = {Jul},
  publisher = {American Physical Society},
  doi = {10.1103/PhysRevLett.127.038301},
  url = {https://link.aps.org/doi/10.1103/PhysRevLett.127.038301}
}

@article{van_tegelen_neural_2025,
	title = {Neural ordinary differential equations for learning and extrapolating system dynamics across bifurcations},
	volume = {35},
	issn = {1054-1500},
	url = {https://doi.org/10.1063/5.0288264},
	doi = {10.1063/5.0288264},
	number = {10},
	urldate = {2025-11-12},
	journal = {Chaos: An Interdisciplinary Journal of Nonlinear Science},
	author = {van Tegelen, Eva and van Voorn, George and Athanasiadis, Ioannis N. and van Heijster, Peter},
	month = oct,
	year = {2025},
	pages = {101103},
}

@article{errico_what_1997,
	title = {What Is an Adjoint Model?},
	issn = {1520-0477},
	url = {https://journals.ametsoc.org/view/journals/bams/78/11/1520-0477_1997_078_2577_wiaam_2_0_co_2.xml},
	journal = {Bulletin of the American Meteorological Society},
	volume = {78},
	issue = {11},
	
	urldate = {2025-05-16},
	author = {Errico, Ronald M.},
	month = nov,
	year = {1997},
	note = {Section: Bulletin of the American Meteorological Society},
}

@article{bradley_nonlinear_2015,
	title = {Nonlinear time-series analysis revisited},
	volume = {25},
	issn = {1054-1500},
	url = {https://doi.org/10.1063/1.4917289},
	doi = {10.1063/1.4917289},
	number = {9},
	urldate = {2025-05-16},
	journal = {Chaos: An Interdisciplinary Journal of Nonlinear Science},
	author = {Bradley, Elizabeth and Kantz, Holger},
	month = apr,
	year = {2015},
	pages = {097610},
}

@book{petersen_ergodic_1989,
	title = {Ergodic {Theory}},
	isbn = {978-0-521-38997-6},
	series = {Cambridge Studies in Advanced Mathematics},
	number = {2},
	doi = {10.1017/CBO9780511608728},
	publisher = {Cambridge University Press},
	address = {Cambridge},
	author = {Petersen, Karl E.},
	month = may,
	year = {1983},
}

@article{eckmann_ergodic_1985,
	title = {Ergodic theory of chaos and strange attractors},
	volume = {57},
	url = {https://link.aps.org/doi/10.1103/RevModPhys.57.617},
	doi = {10.1103/RevModPhys.57.617},
	number = {3},
	urldate = {2025-05-16},
	journal = {Reviews of Modern Physics},
	author = {Eckmann, J. -P. and Ruelle, D.},
	month = jul,
	year = {1985},
	note = {Publisher: American Physical Society},
	pages = {617--656},
}

@article{ni_sensitivity_2019,
	title = {Sensitivity analysis on chaotic dynamical systems by {Finite} {Difference} {Non}-{Intrusive} {Least} {Squares} {Shadowing} ({FD}-{NILSS})},
	volume = {394},
	issn = {0021-9991},
	url = {https://www.sciencedirect.com/science/article/pii/S0021999119304115},
	doi = {10.1016/j.jcp.2019.06.004},
	urldate = {2025-05-16},
	journal = {Journal of Computational Physics},
	author = {Ni, Angxiu and Wang, Qiqi and Fernández, Pablo and Talnikar, Chaitanya},
	month = oct,
	year = {2019},
	pages = {615--631},
}

@article{chater_least_2017,
	title = {Least {Squares} {Shadowing} {Method} for {Sensitivity} {Analysis} of {Differential} {Equations}},
	volume = {55},
	issn = {0036-1429},
	url = {https://epubs.siam.org/doi/abs/10.1137/15M1039067},
	doi = {10.1137/15M1039067},
	number = {6},
	urldate = {2025-05-16},
	journal = {SIAM Journal on Numerical Analysis},
	author = {Chater, Mario and Ni, Angxiu and Blonigan, Patrick J. and Wang, Qiqi},
	month = jan,
	year = {2017},
	note = {Publisher: Society for Industrial and Applied Mathematics},
	pages = {3030--3046},
}

@article{wang_least_2014,
	title = {Least {Squares} {Shadowing} sensitivity analysis of chaotic limit cycle oscillations},
	volume = {267},
	issn = {0021-9991},
	url = {https://www.sciencedirect.com/science/article/pii/S0021999114001715},
	doi = {10.1016/j.jcp.2014.03.002},
	urldate = {2025-05-16},
	journal = {Journal of Computational Physics},
	author = {Wang, Qiqi and Hu, Rui and Blonigan, Patrick},
	month = jun,
	year = {2014},
	pages = {210--224},
}

@article{ni_adjoint_2019,
	title = {Adjoint sensitivity analysis on chaotic dynamical systems by {Non}-{Intrusive} {Least} {Squares} {Adjoint} {Shadowing} ({NILSAS})},
	volume = {395},
	issn = {0021-9991},
	url = {https://www.sciencedirect.com/science/article/pii/S0021999119304437},
	doi = {10.1016/j.jcp.2019.06.035},
	urldate = {2025-05-16},
	journal = {Journal of Computational Physics},
	author = {Ni, Angxiu and Talnikar, Chaitanya},
	month = oct,
	year = {2019},
	pages = {690--709},
}

@article{kochkov2024,
	title = {Neural general circulation models for weather and climate},
	volume = {632},
	copyright = {2024 The Author(s)},
	issn = {1476-4687},
	url = {https://www.nature.com/articles/s41586-024-07744-y},
	doi = {10.1038/s41586-024-07744-y},
	number = {8027},
	urldate = {2025-04-09},
	journal = {Nature},
	author = {Kochkov, Dmitrii and Yuval, Janni and Langmore, Ian and Norgaard, Peter and Smith, Jamie and Mooers, Griffin and Klöwer, Milan and Lottes, James and Rasp, Stephan and Düben, Peter and Hatfield, Sam and Battaglia, Peter and Sanchez-Gonzalez, Alvaro and Willson, Matthew and Brenner, Michael P. and Hoyer, Stephan},
	month = aug,
	year = {2024},
	note = {Publisher: Nature Publishing Group},
	pages = {1060--1066},
}

@article{chen2022,
	title = {Exact mean-field models for spiking neural networks with adaptation},
	volume = {50},
	issn = {1573-6873},
	doi = {10.1007/s10827-022-00825-9},
	number = {4},
	journal = {Journal of Computational Neuroscience},
	author = {Chen, Liang and Campbell, Sue Ann},
	month = nov,
	year = {2022},
	pmid = {35834100},
	pages = {445--469},
}

@article{brunton2016,
	title = {Discovering governing equations from data by sparse identification of nonlinear dynamical systems},
	volume = {113},
	copyright = {©  . Freely available online through the PNAS open access option.},
	issn = {0027-8424, 1091-6490},
	url = {https://www.pnas.org/content/113/15/3932},
	doi = {10.1073/pnas.1517384113},
	number = {15},
	urldate = {2021-04-17},
	journal = {Proceedings of the National Academy of Sciences},
	author = {Brunton, Steven L. and Proctor, Joshua L. and Kutz, J. Nathan},
	month = apr,
	year = {2016},
	pages = {3932--3937},
}

@misc{boyd2010,
	title = {Distributed {Optimization} and {Statistical} {Learning} via the {Alternating} {Direction} {Method} of {Multipliers}},
	url = {http://dx.doi.org/10.1561/2200000016},
	publisher = {Now Publishers},
	author = {Boyd, Stephen},
	year = {2010},
	doi = {10.1561/2200000016},
	note = {Issue: 1
Pages: 1–122
Publication Title: Foundations and Trends® in Machine Learning
Volume: 3},
}

@article{bhagavan2024,
	title = {{DataInterpolations}.jl: {Fast} {Interpolations} of {1D} data},
	volume = {9},
	issn = {2475-9066},
	shorttitle = {{DataInterpolations}.jl},
	url = {https://joss.theoj.org/papers/10.21105/joss.06917},
	doi = {10.21105/joss.06917},
	number = {101},
	urldate = {2025-04-07},
	journal = {Journal of Open Source Software},
	author = {Bhagavan, Sathvik and Koning, Bart de and Maddhashiya, Shubham and Rackauckas, Christopher},
	month = sep,
	year = {2024},
	pages = {6917},
}

@article{balcioglu2023,
	title = {Mapping thalamic innervation to individual {L2}/3 pyramidal neurons and modeling their ‘readout’ of visual input},
	volume = {26},
	copyright = {2023 The Author(s), under exclusive licence to Springer Nature America, Inc.},
	issn = {1546-1726},
	url = {https://www.nature.com/articles/s41593-022-01253-9},
	doi = {10.1038/s41593-022-01253-9},
	number = {3},
	urldate = {2025-05-07},
	journal = {Nature Neuroscience},
	author = {Balcioglu, Aygul and Gillani, Rebecca and Doron, Michael and Burnell, Kendyll and Ku, Taeyun and Erisir, Alev and Chung, Kwanghun and Segev, Idan and Nedivi, Elly},
	month = mar,
	year = {2023},
	note = {Publisher: Nature Publishing Group},
	pages = {470--480},
}

@article{antal2024,
	title = {Achieving {Occam}'s razor: {Deep} learning for optimal model reduction},
	volume = {20},
	issn = {1553-7358},
	shorttitle = {Achieving {Occam}’s razor},
	url = {https://journals.plos.org/ploscompbiol/article?id=10.1371/journal.pcbi.1012283},
	doi = {10.1371/journal.pcbi.1012283},
	number = {7},
	urldate = {2024-07-25},
	journal = {PLOS Computational Biology},
	author = {Antal, Botond B. and Chesebro, Anthony G. and Strey, Helmut H. and Mujica-Parodi, Lilianne R. and Weistuch, Corey},
	month = jul,
	year = {2024},
	note = {Publisher: Public Library of Science},
	pages = {e1012283},
}

@article{ziepke2022,
	title = {Multi-scale organization in communicating active matter},
	volume = {13},
	copyright = {2022 The Author(s)},
	issn = {2041-1723},
	url = {https://www.nature.com/articles/s41467-022-34484-2},
	doi = {10.1038/s41467-022-34484-2},
	
	number = {1},
	urldate = {2025-05-13},
	journal = {Nature Communications},
	author = {Ziepke, Alexander and Maryshev, Ivan and Aranson, Igor S. and Frey, Erwin},
	month = nov,
	year = {2022},
	note = {Publisher: Nature Publishing Group},
	pages = {6727},
}

@article{beregi2023,
	title = {Using scientific machine learning for experimental bifurcation analysis of dynamic systems},
	volume = {184},
	issn = {0888-3270},
	url = {https://www.sciencedirect.com/science/article/pii/S0888327022007348},
	doi = {10.1016/j.ymssp.2022.109649},
	urldate = {2025-05-13},
	journal = {Mechanical Systems and Signal Processing},
	author = {Beregi, Sandor and Barton, David A. W. and Rezgui, Djamel and Neild, Simon},
	month = feb,
	year = {2023},
	pages = {109649},
}

@article{foster2020,
	title = {A Bayesian Approach to Regional Decadal Predictability: Sparse Parameter Estimation in High-Dimensional Linear Inverse Models of High-Latitude Sea Surface Temperature Variability},
	volume = {},
	journal = {Journal of Climate},
	url = {https://journals.ametsoc.org/view/journals/clim/33/14/jcliD190769.xml},
	doi = {10.1175/JCLI-D-19-0769.1},
	urldate = {2025-05-13},
	author = {Foster, Dallas and Comeau, Darin and Urban, Nathan M.},
	month = jul,
	year = {2020},
	note = {Section: Journal of Climate},
}

@article{fruengel2025,
	title = {Sparse connectivity enables efficient information processing in cortex-like artificial neural networks},
	volume = {19},
	issn = {1662-5110},
	url = {https://www.frontiersin.orghttps://www.frontiersin.org/journals/neural-circuits/articles/10.3389/fncir.2025.1528309/full},
	doi = {10.3389/fncir.2025.1528309},
	
	urldate = {2025-05-07},
	journal = {Frontiers in Neural Circuits},
	author = {Fruengel, Rieke and Oberlaender, Marcel},
	month = mar,
	year = {2025},
	note = {Publisher: Frontiers},
	pages = {1528309},
}

@article{feinerman2018,
	title = {The physics of cooperative transport in groups of ants},
	volume = {14},
	copyright = {2018 Springer Nature Limited},
	issn = {1745-2481},
	url = {https://www.nature.com/articles/s41567-018-0107-y},
	doi = {10.1038/s41567-018-0107-y},
	
	number = {7},
	urldate = {2025-05-06},
	journal = {Nature Physics},
	author = {Feinerman, Ofer and Pinkoviezky, Itai and Gelblum, Aviram and Fonio, Ehud and Gov, Nir S.},
	month = jul,
	year = {2018},
	note = {Publisher: Nature Publishing Group},
	pages = {683--693},
}

@article{nozari2024,
	title = {Macroscopic resting-state brain dynamics are best described by linear models},
	volume = {8},
	copyright = {2023 The Author(s)},
	issn = {2157-846X},
	url = {https://www.nature.com/articles/s41551-023-01117-y},
	doi = {10.1038/s41551-023-01117-y},
	
	number = {1},
	urldate = {2025-05-06},
	journal = {Nature Biomedical Engineering},
	author = {Nozari, Erfan and Bertolero, Maxwell A. and Stiso, Jennifer and Caciagli, Lorenzo and Cornblath, Eli J. and He, Xiaosong and Mahadevan, Arun S. and Pappas, George J. and Bassett, Dani S.},
	month = jan,
	year = {2024},
	note = {Publisher: Nature Publishing Group},
	pages = {68--84},
}

@article{stoica1989,
	title = {On multistep prediction error methods for time series models},
	volume = {8},
	copyright = {Copyright © 1989 John Wiley \& Sons, Ltd.},
	issn = {1099-131X},
	url = {https://onlinelibrary.wiley.com/doi/abs/10.1002/for.3980080402},
	doi = {10.1002/for.3980080402},
	
	number = {4},
	urldate = {2024-12-19},
	journal = {Journal of Forecasting},
	author = {Stoica, Petre},
	year = {1989},
	pages = {357--368},
}

@article{halgren2019,
	title = {The generation and propagation of the human alpha rhythm},
	volume = {116},
	issn = {0027-8424},
	url = {https://www.ncbi.nlm.nih.gov/pmc/articles/PMC6876194/},
	doi = {10.1073/pnas.1913092116},
	number = {47},
	urldate = {2025-05-06},
	journal = {Proceedings of the National Academy of Sciences of the United States of America},
	author = {Halgren, Mila and Ulbert, István and Bastuji, Hélène and Fabó, Dániel and Erőss, Lorand and Rey, Marc and Devinsky, Orrin and Doyle, Werner K. and Mak-McCully, Rachel and Halgren, Eric and Wittner, Lucia and Chauvel, Patrick and Heit, Gary and Eskandar, Emad and Mandell, Arnold and Cash, Sydney S.},
	month = nov,
	year = {2019},
	pmid = {31685634},
	pmcid = {PMC6876194},
	pages = {23772--23782},
}

@article{oleary2022,
	title = {Stochastic physics-informed neural ordinary differential equations},
	volume = {468},
	issn = {0021-9991},
	url = {https://www.sciencedirect.com/science/article/pii/S0021999122005289},
	doi = {10.1016/j.jcp.2022.111466},
	urldate = {2025-05-06},
	journal = {Journal of Computational Physics},
	author = {O'Leary, Jared and Paulson, Joel A. and Mesbah, Ali},
	month = nov,
	year = {2022},
	pages = {111466},
}

@book{kuramoto1984,
	address = {Berlin, Heidelberg},
	series = {Springer {Series} in {Synergetics}},
	title = {Chemical {Oscillations}, {Waves}, and {Turbulence}},
	volume = {19},
	copyright = {http://www.springer.com/tdm},
	isbn = {978-3-642-69691-6 978-3-642-69689-3},
	url = {http://link.springer.com/10.1007/978-3-642-69689-3},
	urldate = {2025-05-05},
	publisher = {Springer},
	author = {Kuramoto, Yoshiki},
	editor = {Haken, Hermann},
	year = {1984},
	doi = {10.1007/978-3-642-69689-3},
}

@article{mormann2000,
	title = {Mean phase coherence as a measure for phase synchronization and its application to the {EEG} of epilepsy patients},
	volume = {144},
	issn = {0167-2789},
	url = {https://www.sciencedirect.com/science/article/pii/S0167278900000877},
	doi = {10.1016/S0167-2789(00)00087-7},
	number = {3},
	urldate = {2025-05-05},
	journal = {Physica D: Nonlinear Phenomena},
	author = {Mormann, Florian and Lehnertz, Klaus and David, Peter and E. Elger, Christian},
	month = oct,
	year = {2000},
	pages = {358--369},
}

@misc{peyrache2018,
	title = {Activity of neurons in rat medial prefrontal cortex during learning and sleep.},
	doi = {dx.doi.org/10.6080/K0KH0KH5},
	publisher = {CRCNS.org},
	author = {Peyrache, Adrien and Khamassi, Mehdi and Benchenane, Karim and Wiener, Sidney and Battaglia, Francesco},
	year = {2018},
}

@article{peyrache2009,
	title = {Replay of rule-learning related neural patterns in the prefrontal cortex during sleep},
	volume = {12},
	issn = {1546-1726},
	doi = {10.1038/nn.2337},
	
	number = {7},
	journal = {Nature Neuroscience},
	author = {Peyrache, Adrien and Khamassi, Mehdi and Benchenane, Karim and Wiener, Sidney I. and Battaglia, Francesco P.},
	month = jul,
	year = {2009},
	pmid = {19483687},
	pages = {919--926},
}

@article{saez2018,
	title = {Encoding of {Multiple} {Reward}-{Related} {Computations} in {Transient} and {Sustained} {High}-{Frequency} {Activity} in {Human} {OFC}},
	volume = {28},
	issn = {0960-9822},
	url = {https://www.cell.com/current-biology/abstract/S0960-9822(18)30975-8},
	doi = {10.1016/j.cub.2018.07.045},
	
	number = {18},
	urldate = {2025-04-30},
	journal = {Current Biology},
	author = {Saez, Ignacio and Lin, Jack and Stolk, Arjen and Chang, Edward and Parvizi, Josef and Schalk, Gerwin and Knight, Robert T. and Hsu, Ming},
	month = sep,
	year = {2018},
	pmid = {30220499},
	note = {Publisher: Elsevier},
	pages = {2889--2899.e3},
}

@misc{saez2018a,
	title = {High-frequency activity of human orbitofrontal sites during decision-making play.},
	doi = {dx.doi.org/10.6080/K0VM49GF},
	publisher = {CRCNS.org},
	author = {Saez, Ignacio and Lin, Jack and Stolk, Arjen and Chang, Edward and Parvizi, Josef and Schalk, Gerwin and Knight, Robert and Hsu, Ming},
	year = {2018},
}

@article{hernandez-perez2020,
	title = {Medial entorhinal cortex activates in a traveling wave in the rat},
	volume = {9},
	issn = {2050-084X},
	doi = {10.7554/eLife.52289},
	
	journal = {eLife},
	author = {Hernández-Pérez, J. Jesús and Cooper, Keiland W. and Newman, Ehren L.},
	month = feb,
	year = {2020},
	pmid = {32057292},
	pmcid = {PMC7046467},
	pages = {e52289},
}

@misc{hernandez-perez2020a,
	title = {Extracellular recordings from across the dorsoventral axis of the medial entorhinal cortex of the rat.},
	doi = {dx.doi.org/10.6080/K0C53J2R},
	publisher = {CRCNS.org},
	author = {Hernández-Pérez, JJ and Cooper, KW and Newman, EL},
	year = {2020},
}

@misc{juliusmartensen2021,
	title = {{DataDrivenDiffEq}.jl},
	url = {https://doi.org/10.5281/zenodo.5083412},
	publisher = {Zenodo},
	author = {{Martensen, Julius} and Rackauckas, Christopher and {others}},
	month = jul,
	year = {2021},
	doi = {10.5281/zenodo.5083412},
}

@book{sprott2022,
	title = {Elegant {Circuits}: {Simple} {Chaotic} {Oscillators}},
	isbn = {9789811239991 9789811240003},
	shorttitle = {Elegant {Circuits}},
	url = {https://www.worldscientific.com/worldscibooks/10.1142/12362},
	
	urldate = {2025-04-24},
	publisher = {World Scientific},
	author = {Sprott, Julien Clinton and Thio, Wesley Joo-Chen},
	month = feb,
	year = {2022},
	doi = {10.1142/12362},
}

@article{petrzela2019,
	title = {Minimal {Realizations} of {Autonomous} {Chaotic} {Oscillators} {Based} on {Trans}-{Immittance} {Filters}},
	volume = {7},
	issn = {2169-3536},
	url = {https://ieeexplore.ieee.org/document/8630955},
	doi = {10.1109/ACCESS.2019.2896656},
	urldate = {2025-04-24},
	journal = {IEEE Access},
	author = {Petrzela, Jiri and Polak, Ladislav},
	year = {2019},
	pages = {17561--17577},
}

@article{kuznetsov2023,
	title = {Hidden attractors in {Chua} circuit: mathematical theory meets physical experiments},
	volume = {111},
	issn = {1573-269X},
	shorttitle = {Hidden attractors in {Chua} circuit},
	url = {https://doi.org/10.1007/s11071-022-08078-y},
	doi = {10.1007/s11071-022-08078-y},
	
	number = {6},
	urldate = {2025-04-16},
	journal = {Nonlinear Dynamics},
	author = {Kuznetsov, Nikolay and Mokaev, Timur and Ponomarenko, Vladimir and Seleznev, Evgeniy and Stankevich, Nataliya and Chua, Leon},
	month = mar,
	year = {2023},
	pages = {5859--5887},
}

@article{graf2024,
	title = {A bifurcation integrates information from many noisy ion channels and allows for milli-{Kelvin} thermal sensitivity in the snake pit organ},
	volume = {121},
	url = {https://www.pnas.org/doi/10.1073/pnas.2308215121},
	doi = {10.1073/pnas.2308215121},
	number = {6},
	urldate = {2025-04-09},
	journal = {Proceedings of the National Academy of Sciences},
	author = {Graf, Isabella R. and Machta, Benjamin B.},
	month = feb,
	year = {2024},
	note = {Publisher: Proceedings of the National Academy of Sciences},
	pages = {e2308215121},
}

@article{tian2022,
	title = {Theoretical foundations of studying criticality in the brain},
	volume = {6},
	issn = {2472-1751},
	url = {https://doi.org/10.1162/netn_a_00269},
	doi = {10.1162/netn_a_00269},
	number = {4},
	urldate = {2025-04-09},
	journal = {Network Neuroscience},
	author = {Tian, Yang and Tan, Zeren and Hou, Hedong and Li, Guoqi and Cheng, Aohua and Qiu, Yike and Weng, Kangyu and Chen, Chun and Sun, Pei},
	month = oct,
	year = {2022},
	pages = {1148--1185},
}

@article{habibollahi2023,
	title = {Critical dynamics arise during structured information presentation within embodied in vitro neuronal networks},
	volume = {14},
	copyright = {2023 The Author(s)},
	issn = {2041-1723},
	url = {https://www.nature.com/articles/s41467-023-41020-3},
	doi = {10.1038/s41467-023-41020-3},
	
	number = {1},
	urldate = {2025-04-09},
	journal = {Nature Communications},
	author = {Habibollahi, Forough and Kagan, Brett J. and Burkitt, Anthony N. and French, Chris},
	month = aug,
	year = {2023},
	note = {Publisher: Nature Publishing Group},
	pages = {5287},
}

@article{bakarji2023a,
	title = {Discovering governing equations from partial measurements with deep delay autoencoders},
	volume = {479},
	url = {https://royalsocietypublishing.org/doi/10.1098/rspa.2023.0422},
	doi = {10.1098/rspa.2023.0422},
	number = {2276},
	urldate = {2025-03-18},
	journal = {Proceedings of the Royal Society A: Mathematical, Physical and Engineering Sciences},
	author = {Bakarji, Joseph and Champion, Kathleen and Nathan Kutz, J. and Brunton, Steven L.},
	month = aug,
	year = {2023},
	note = {Publisher: Royal Society},
	pages = {20230422},
}

@article{pagan2019,
	title = {Game theoretical inference of human behavior in social networks},
	volume = {10},
	copyright = {2019 The Author(s)},
	issn = {2041-1723},
	url = {https://www.nature.com/articles/s41467-019-13148-8},
	doi = {10.1038/s41467-019-13148-8},
	
	number = {1},
	urldate = {2025-04-09},
	journal = {Nature Communications},
	author = {Pagan, Nicolò and Dörfler, Florian},
	month = dec,
	year = {2019},
	note = {Publisher: Nature Publishing Group},
	pages = {5507},
}

@article{larsson2009,
	series = {15th {IFAC} {Symposium} on {System} {Identification}},
	title = {Direct prediction-error identification of unstable nonlinear systems applied to flight test data},
	volume = {42},
	issn = {1474-6670},
	url = {https://www.sciencedirect.com/science/article/pii/S1474667016386384},
	doi = {10.3182/20090706-3-FR-2004.00024},
	number = {10},
	urldate = {2024-12-19},
	journal = {IFAC Proceedings Volumes},
	author = {Larsson, Roger and Sjanic, Zoran and Enqvist, Martin and Ljung, Lennart},
	month = jan,
	year = {2009},
	pages = {144--149},
}

@article{pillonetto2025,
	title = {Deep networks for system identification: {A} survey},
	volume = {171},
	issn = {0005-1098},
	shorttitle = {Deep networks for system identification},
	url = {https://www.sciencedirect.com/science/article/pii/S0005109824004011},
	doi = {10.1016/j.automatica.2024.111907},
	urldate = {2024-12-19},
	journal = {Automatica},
	author = {Pillonetto, Gianluigi and Aravkin, Aleksandr and Gedon, Daniel and Ljung, Lennart and Ribeiro, Antônio H. and Schön, Thomas B.},
	month = jan,
	year = {2025},
	pages = {111907},
}

@inproceedings{sun2021,
	title = {Nonlinear {System} {Identification}: {Prediction} {Error} {Method} vs {Neural} {Network}},
	shorttitle = {Nonlinear {System} {Identification}},
	url = {https://ieeexplore.ieee.org/document/9493336},
	doi = {10.1109/MOCAST52088.2021.9493336},
	urldate = {2024-12-19},
	booktitle = {2021 10th {International} {Conference} on {Modern} {Circuits} and {Systems} {Technologies} ({MOCAST})},
	author = {Sun, Jinming and Huang, Yanqiu and Yu, Wanli and Garcia-Ortiz, Alberto},
	month = jul,
	year = {2021},
	pages = {1--4},
}

@article{goldobin2024,
	title = {Discrete {Synaptic} {Events} {Induce} {Global} {Oscillations} in {Balanced} {Neural} {Networks}},
	volume = {133},
	url = {https://link.aps.org/doi/10.1103/PhysRevLett.133.238401},
	doi = {10.1103/PhysRevLett.133.238401},
	number = {23},
	urldate = {2025-02-04},
	journal = {Physical Review Letters},
	author = {Goldobin, Denis S. and di Volo, Matteo and Torcini, Alessandro},
	month = dec,
	year = {2024},
	note = {Publisher: American Physical Society},
	pages = {238401},
}

@article{montbrio2015,
	title = {Macroscopic {Description} for {Networks} of {Spiking} {Neurons}},
	volume = {5},
	url = {https://link.aps.org/doi/10.1103/PhysRevX.5.021028},
	doi = {10.1103/PhysRevX.5.021028},
	number = {2},
	urldate = {2025-02-07},
	journal = {Physical Review X},
	author = {Montbrió, Ernest and Pazó, Diego and Roxin, Alex},
	month = jun,
	year = {2015},
	note = {Publisher: American Physical Society},
	pages = {021028},
}

@misc{cranmer2023,
	title = {Interpretable {Machine} {Learning} for {Science} with {PySR} and {SymbolicRegression}.jl},
	url = {http://arxiv.org/abs/2305.01582},
	doi = {10.48550/arXiv.2305.01582},
	eprint = {2305.01582},
	archivePrefix = {arXiv},
	primaryClass = {astro-ph.IM},
	urldate = {2023-08-24},
	publisher = {arXiv},
	author = {Cranmer, Miles},
	month = may,
	year = {2023},
}

@article{ott2008a,
	title = {Low dimensional behavior of large systems of globally coupled oscillators},
	volume = {18},
	issn = {1054-1500, 1089-7682},
	url = {http://aip.scitation.org/doi/10.1063/1.2930766},
	doi = {10.1063/1.2930766},
	
	number = {3},
	urldate = {2022-01-10},
	journal = {Chaos: An Interdisciplinary Journal of Nonlinear Science},
	author = {Ott, Edward and Antonsen, Thomas M.},
	month = sep,
	year = {2008},
	pages = {037113},
}

@article{tibshirani1996,
	title = {Regression {Shrinkage} and {Selection} via the {Lasso}},
	volume = {58},
	issn = {0035-9246},
	url = {https://www.jstor.org/stable/2346178},
	number = {1},
	urldate = {2025-03-14},
	journal = {Journal of the Royal Statistical Society. Series B (Methodological)},
	author = {Tibshirani, Robert},
	year = {1996},
	note = {Publisher: [Royal Statistical Society, Oxford University Press]},
	pages = {267--288},
}

@article{goncalves2020,
	title = {Training deep neural density estimators to identify mechanistic models of neural dynamics},
	volume = {9},
	issn = {2050-084X},
	url = {https://elifesciences.org/articles/56261},
	doi = {10.7554/eLife.56261},
	
	urldate = {2023-06-19},
	journal = {eLife},
	author = {Goncalves, Pedro J and Lueckmann, Jan-Matthis and Deistler, Michael and Nonnenmacher, Marcel and Öcal, Kaan and Bassetto, Giacomo and Chintaluri, Chaitanya and Podlaski, William F and Haddad, Sara A and Vogels, Tim P and Greenberg, David S and Macke, Jakob H},
	month = sep,
	year = {2020},
	pages = {e56261},
}

@article{divolo2018,
	title = {Transition from {Asynchronous} to {Oscillatory} {Dynamics} in {Balanced} {Spiking} {Networks} with {Instantaneous} {Synapses}},
	volume = {121},
	url = {https://link.aps.org/doi/10.1103/PhysRevLett.121.128301},
	doi = {10.1103/PhysRevLett.121.128301},
	number = {12},
	urldate = {2024-02-11},
	journal = {Physical Review Letters},
	author = {di Volo, Matteo and Torcini, Alessandro},
	month = sep,
	year = {2018},
	note = {Publisher: American Physical Society},
	pages = {128301},
}

@article{park1991,
	title = {Universal {Approximation} {Using} {Radial}-{Basis}-{Function} {Networks}},
	volume = {3},
	issn = {0899-7667},
	url = {https://ieeexplore.ieee.org/document/6797088},
	doi = {10.1162/neco.1991.3.2.246},
	number = {2},
	urldate = {2025-02-25},
	journal = {Neural Computation},
	author = {Park, J. and Sandberg, I. W.},
	month = jun,
	year = {1991},
	note = {Conference Name: Neural Computation},
	pages = {246--257},
}

@misc{rackauckas2021,
	title = {Universal {Differential} {Equations} for {Scientific} {Machine} {Learning}},
	url = {https://arxiv.org/abs/2001.04385},
	doi = {10.48550/arXiv.2001.04385},
	eprint = {2001.04385},
	archivePrefix = {arXiv},
	primaryClass = {cs.LG},
	
	urldate = {2022-02-28},
	author = {Rackauckas, Christopher and Ma, Yingbo and Martensen, Julius and Warner, Collin and Zubov, Kirill and Supekar, Rohit and Skinner, Dominic and Ramadhan, Ali and Edelman, Alan},
	month = jan,
	year = {2020},
}

@article{ji2021,
	title = {Stiff-{PINN}: {Physics}-{Informed} {Neural} {Network} for {Stiff} {Chemical} {Kinetics}},
	volume = {125},
	issn = {1089-5639},
	shorttitle = {Stiff-{PINN}},
	url = {https://doi.org/10.1021/acs.jpca.1c05102},
	doi = {10.1021/acs.jpca.1c05102},
	number = {36},
	urldate = {2025-02-03},
	journal = {The Journal of Physical Chemistry A},
	author = {Ji, Weiqi and Qiu, Weilun and Shi, Zhiyu and Pan, Shaowu and Deng, Sili},
	month = sep,
	year = {2021},
	note = {Publisher: American Chemical Society},
	pages = {8098--8106},
}

@article{kim2021,
	title = {Stiff neural ordinary differential equations},
	volume = {31},
	issn = {1054-1500},
	url = {https://doi.org/10.1063/5.0060697},
	doi = {10.1063/5.0060697},
	number = {9},
	urldate = {2023-08-25},
	journal = {Chaos: An Interdisciplinary Journal of Nonlinear Science},
	author = {Kim, Suyong and Ji, Weiqi and Deng, Sili and Ma, Yingbo and Rackauckas, Christopher},
	month = sep,
	year = {2021},
	pages = {093122},
}

@article{ding2010,
	title = {Thalamic {Gating} of {Corticostriatal} {Signaling} by {Cholinergic} {Interneurons}},
	volume = {67},
	issn = {08966273},
	url = {https://linkinghub.elsevier.com/retrieve/pii/S0896627310004757},
	doi = {10.1016/j.neuron.2010.06.017},
	
	number = {2},
	urldate = {2023-09-11},
	journal = {Neuron},
	author = {Ding, Jun B. and Guzman, Jaime N. and Peterson, Jayms D. and Goldberg, Joshua A. and Surmeier, D. James},
	month = jul,
	year = {2010},
	pages = {294--307},
}

@article{king2009,
	title = {The automation of science},
	volume = {324},
	number = {5923},
	journal = {Science},
	author = {King, Ross D and Rowland, Jem and Oliver, Stephen G and Young, Michael and Aubrey, Wayne and Byrne, Emma and Liakata, Maria and Markham, Magdalena and Pir, Pınar and Soldatova, Larisa N and {others}},
	year = {2009},
	note = {Publisher: American Association for the Advancement of Science},
	pages = {85--89},
}

@article{ljung2002,
	title = {Prediction error estimation methods},
	volume = {21},
	issn = {1531-5878},
	url = {https://doi.org/10.1007/BF01211648},
	doi = {10.1007/BF01211648},
	
	number = {1},
	urldate = {2025-02-20},
	journal = {Circuits, Systems and Signal Processing},
	author = {Ljung, Lennart},
	month = jan,
	year = {2002},
	pages = {11--21},
}

@incollection{coombes2019,
	address = {Cham},
	title = {Next {Generation} {Neural} {Mass} {Models}},
	isbn = {978-3-319-71047-1 978-3-319-71048-8},
	url = {http://link.springer.com/10.1007/978-3-319-71048-8_1},
	
	urldate = {2022-01-10},
	booktitle = {Nonlinear {Dynamics} in {Computational} {Neuroscience}},
	publisher = {Springer International Publishing},
	author = {Coombes, Stephen and Byrne, Aine},
	editor = {Corinto, Fernando and Torcini, Alessandro},
	year = {2019},
	pages = {1--16},
}

@book{datseris2022,
	address = {Cham, Switzerland},
	title = {Nonlinear dynamics: {A} concise introduction interlaced with code},
	url = {https://doi.org/10.1007/978-3-030-91032-7},
	
	publisher = {Springer Nature},
	author = {Datseris, George and Parlitz, Ulrich},
	year = {2022},
	doi = {10.1007/978-3-030-91032-7},
}

@article{chandramoorthy2021,
	title = {On the probability of finding nonphysical solutions through shadowing},
	volume = {440},
	issn = {0021-9991},
	url = {https://www.sciencedirect.com/science/article/pii/S0021999121002849},
	doi = {10.1016/j.jcp.2021.110389},
	urldate = {2025-02-07},
	journal = {Journal of Computational Physics},
	author = {Chandramoorthy, Nisha and Wang, Qiqi},
	month = sep,
	year = {2021},
	pages = {110389},
}

@article{goldobin2019,
	title = {Ott-{Antonsen} ansatz truncation of a circular cumulant series},
	volume = {1},
	issn = {2643-1564},
	url = {https://link.aps.org/doi/10.1103/PhysRevResearch.1.033139},
	doi = {10.1103/PhysRevResearch.1.033139},
	
	number = {3},
	urldate = {2024-05-23},
	journal = {Physical Review Research},
	author = {Goldobin, Denis S. and Dolmatova, Anastasiya V.},
	month = dec,
	year = {2019},
	pages = {033139},
}

@inproceedings{craddock1996,
	title = {Multi-layer radial basis function networks. {An} extension to the radial basis function},
	volume = {2},
	url = {https://ieeexplore.ieee.org/document/548981},
	doi = {10.1109/ICNN.1996.548981},
	urldate = {2025-02-25},
	booktitle = {Proceedings of {International} {Conference} on {Neural} {Networks} ({ICNN}'96)},
	author = {Craddock, R.J. and Warwick, K.},
	month = jun,
	year = {1996},
	pages = {700--705 vol.2},
}

@article{zhao2019,
	title = {Multi-layer radial basis function neural network based on multi-scale kernel learning},
	volume = {82},
	issn = {1568-4946},
	url = {https://www.sciencedirect.com/science/article/pii/S1568494619303205},
	doi = {10.1016/j.asoc.2019.105541},
	urldate = {2025-02-25},
	journal = {Applied Soft Computing},
	author = {Zhao, Yang and Pei, Jihong and Chen, Hao},
	month = sep,
	year = {2019},
	pages = {105541},
}

@article{nicola2013,
	title = {Mean-field models for heterogeneous networks of two-dimensional integrate and fire neurons},
	volume = {7},
	issn = {1662-5188},
	doi = {10.3389/fncom.2013.00184},
	journal = {Frontiers in Computational Neuroscience},
	author = {Nicola, Wilten and Campbell, Sue Ann},
	year = {2013},
	pages = {184},
}

@article{weistuch2021,
	title = {Metabolism modulates network synchrony in the aging brain},
	volume = {118},
	issn = {1091-6490 (Electronic) 0027-8424 (Print) 0027-8424 (Linking)},
	url = {https://www.ncbi.nlm.nih.gov/pubmed/34588302},
	doi = {10.1073/pnas.2025727118},
	number = {40},
	journal = {Proc Natl Acad Sci U S A},
	author = {Weistuch, C. and Mujica-Parodi, L. R. and Razban, R. M. and Antal, B. and van Nieuwenhuizen, H. and Amgalan, A. and Dill, K. A.},
	month = oct,
	year = {2021},
}

@article{roberts2019,
	title = {Metastable brain waves},
	volume = {10},
	copyright = {2019 The Author(s)},
	issn = {2041-1723},
	url = {https://www.nature.com/articles/s41467-019-08999-0},
	doi = {10.1038/s41467-019-08999-0},
	
	number = {1},
	urldate = {2025-02-24},
	journal = {Nature Communications},
	author = {Roberts, James A. and Gollo, Leonardo L. and Abeysuriya, Romesh G. and Roberts, Gloria and Mitchell, Philip B. and Woolrich, Mark W. and Breakspear, Michael},
	month = mar,
	year = {2019},
	note = {Publisher: Nature Publishing Group},
	pages = {1056},
}

@article{breakspear2003,
	title = {Modulation of excitatory synaptic coupling facilitates synchronization and complex dynamics in a biophysical model of neuronal dynamics},
	volume = {14},
	issn = {0954-898X},
	url = {https://doi.org/10.1088/0954-898X_14_4_305},
	doi = {10.1088/0954-898X_14_4_305},
	number = {4},
	urldate = {2022-09-29},
	journal = {Network: Computation in Neural Systems},
	author = {Breakspear, Michael and Terry, John R and Friston, Karl J},
	month = nov,
	year = {2003},
	pmid = {14653499},
	pages = {703--732},
}

@article{hannay2018,
	title = {Macroscopic models for networks of coupled biological oscillators},
	volume = {4},
	url = {https://www.science.org/doi/10.1126/sciadv.1701047},
	doi = {10.1126/sciadv.1701047},
	number = {8},
	urldate = {2025-02-08},
	journal = {Science Advances},
	author = {Hannay, Kevin M. and Forger, Daniel B. and Booth, Victoria},
	month = aug,
	year = {2018},
	note = {Publisher: American Association for the Advancement of Science},
	pages = {e1701047},
}

@article{sheheitli2024,
	title = {Incorporating slow {NMDA}-type receptors with nonlinear voltage-dependent magnesium block in a next generation neural mass model: derivation and dynamics},
	volume = {52},
	issn = {1573-6873},
	shorttitle = {Incorporating slow {NMDA}-type receptors with nonlinear voltage-dependent magnesium block in a next generation neural mass model},
	url = {https://doi.org/10.1007/s10827-024-00874-2},
	doi = {10.1007/s10827-024-00874-2},
	
	number = {3},
	urldate = {2024-09-09},
	journal = {Journal of Computational Neuroscience},
	author = {Sheheitli, Hiba and Jirsa, Viktor},
	month = aug,
	year = {2024},
	pages = {207--222},
}

@article{chesebro2023,
	title = {Ion gradient-driven bifurcations of a multi-scale neuronal model},
	volume = {167},
	issn = {0960-0779},
	url = {https://www.sciencedirect.com/science/article/pii/S0960077923000218},
	doi = {10.1016/j.chaos.2023.113120},
	urldate = {2024-03-20},
	journal = {Chaos, Solitons \& Fractals},
	author = {Chesebro, Anthony G. and Mujica-Parodi, Lilianne R. and Weistuch, Corey},
	month = feb,
	year = {2023},
	pages = {113120},
}

@article{adam2024,
	title = {Ketamine can produce oscillatory dynamics by engaging mechanisms dependent on the kinetics of {NMDA} receptors},
	volume = {121},
	url = {https://www.pnas.org/doi/10.1073/pnas.2402732121},
	doi = {10.1073/pnas.2402732121},
	number = {22},
	urldate = {2024-06-25},
	journal = {Proceedings of the National Academy of Sciences},
	author = {Adam, Elie and Kowalski, Marek and Akeju, Oluwaseun and Miller, Earl K. and Brown, Emery N. and McCarthy, Michelle M. and Kopell, Nancy},
	month = may,
	year = {2024},
	pages = {e2402732121},
}

@article{clusella2024,
	title = {Exact low-dimensional description for fast neural oscillations with low firing rates},
	volume = {109},
	url = {https://link.aps.org/doi/10.1103/PhysRevE.109.014229},
	doi = {10.1103/PhysRevE.109.014229},
	number = {1},
	urldate = {2024-04-05},
	journal = {Physical Review E},
	author = {Clusella, Pau and Montbrió, Ernest},
	month = jan,
	year = {2024},
	note = {Publisher: American Physical Society},
	pages = {014229},
}

@article{wilson1972,
	title = {Excitatory and {Inhibitory} {Interactions} in {Localized} {Populations} of {Model} {Neurons}},
	volume = {12},
	issn = {0006-3495},
	url = {https://www.sciencedirect.com/science/article/pii/S0006349572860685},
	doi = {10.1016/S0006-3495(72)86068-5},
	
	number = {1},
	urldate = {2021-09-08},
	journal = {Biophysical Journal},
	author = {Wilson, Hugh R. and Cowan, Jack D.},
	month = jan,
	year = {1972},
	pages = {1--24},
}

@article{ellis2023,
	title = {{DreamCoder}: growing generalizable, interpretable knowledge with wake–sleep {Bayesian} program learning},
	volume = {381},
	shorttitle = {{DreamCoder}},
	url = {https://royalsocietypublishing.org/doi/10.1098/rsta.2022.0050},
	doi = {10.1098/rsta.2022.0050},
	number = {2251},
	urldate = {2024-02-26},
	journal = {Philosophical Transactions of the Royal Society A: Mathematical, Physical and Engineering Sciences},
	author = {Ellis, Kevin and Wong, Lionel and Nye, Maxwell and Sablé-Meyer, Mathias and Cary, Luc and Anaya Pozo, Lore and Hewitt, Luke and Solar-Lezama, Armando and Tenenbaum, Joshua B.},
	month = jun,
	year = {2023},
	note = {Publisher: Royal Society},
	pages = {20220050},
}

@article{datseris2018,
	title = {{DynamicalSystems}.jl: {A} {Julia} software library for chaos and nonlinear dynamics},
	volume = {3},
	url = {https://doi.org/10.21105/joss.00598},
	doi = {10.21105/joss.00598},
	number = {23},
	journal = {Journal of Open Source Software},
	author = {Datseris, George},
	month = mar,
	year = {2018},
	pages = {598},
}

@article{breakspear2017,
	title = {Dynamic models of large-scale brain activity},
	volume = {20},
	issn = {1546-1726 (Electronic) 1097-6256 (Linking)},
	url = {https://www.ncbi.nlm.nih.gov/pubmed/28230845},
	doi = {10.1038/nn.4497},
	number = {3},
	journal = {Nat Neurosci},
	author = {Breakspear, M.},
	month = feb,
	year = {2017},
	pages = {340--352},
}

@article{jansen1995,
	title = {Electroencephalogram and visual evoked potential generation in a mathematical model of coupled cortical columns},
	volume = {73},
	
	journal = {Biological Cybernetics},
	author = {Jansen, Ben H and Rit, Vincent G},
	year = {1995},
	pages = {357--366},
}

@inproceedings{loshchilov2019,
	title = {Decoupled {Weight} {Decay} {Regularization}},
	url = {https://openreview.net/forum?id=Bkg6RiCqY7},
	booktitle = {International Conference on Learning Representations},
	eprint = {1711.05101},
	archivePrefix = {arXiv},
	primaryClass = {cs.LG},
	urldate = {2025-03-09},
	author = {Loshchilov, Ilya and Hutter, Frank},
	month = may,
	year = {2019},
}

@inproceedings{cranmer2020,
	title = {Discovering {Symbolic} {Models} from {Deep} {Learning} with {Inductive} {Biases}},
	url = {https://proceedings.neurips.cc/paper/2020/hash/c9f2f917078bd2db12f23c3b413d9cba-Abstract.html},
	
	booktitle = {{NeurIPS} 2020},
	author = {Cranmer, Miles and Sanchez-Gonzalez, Alvaro and Battaglia, Peter and Xu, Rui and Cranmer, Kyle and Spergel, David and Ho, Shirley},
	year = {2020},
	pages = {14},
}

@article{schmidt2009,
	title = {Distilling {Free}-{Form} {Natural} {Laws} from {Experimental} {Data}},
	volume = {324},
	url = {https://www.science.org/doi/abs/10.1126/science.1165893},
	doi = {10.1126/science.1165893},
	number = {5923},
	urldate = {2022-02-18},
	journal = {Science},
	author = {Schmidt, Michael and Lipson, Hod},
	month = apr,
	year = {2009},
	note = {Publisher: American Association for the Advancement of Science},
	pages = {81--85},
}

@article{lubenov2009,
	title = {Hippocampal theta oscillations are travelling waves},
	volume = {459},
	copyright = {2009 Macmillan Publishers Limited. All rights reserved},
	issn = {1476-4687},
	url = {https://www.nature.com/articles/nature08010},
	doi = {10.1038/nature08010},
	
	number = {7246},
	urldate = {2025-02-21},
	journal = {Nature},
	author = {Lubenov, Evgueniy V. and Siapas, Athanassios G.},
	month = may,
	year = {2009},
	note = {Publisher: Nature Publishing Group},
	pages = {534--539},
}

@article{schmidt2020,
	title = {Bumps and oscillons in networks of spiking neurons},
	volume = {30},
	issn = {1054-1500, 1089-7682},
	url = {https://aip.scitation.org/doi/10.1063/1.5135579},
	doi = {10.1063/1.5135579},
	
	number = {3},
	urldate = {2022-09-29},
	journal = {Chaos: An Interdisciplinary Journal of Nonlinear Science},
	author = {Schmidt, Helmut and Avitabile, Daniele},
	month = mar,
	year = {2020},
	pages = {033133},
}

@article{nandi2024,
	title = {Bursting gamma oscillations in neural mass models},
	volume = {18},
	issn = {1662-5188},
	url = {https://www.frontiersin.org/journals/computational-neuroscience/articles/10.3389/fncom.2024.1422159/full},
	doi = {10.3389/fncom.2024.1422159},
	
	urldate = {2024-09-09},
	journal = {Frontiers in Computational Neuroscience},
	author = {Nandi, Manoj Kumar and Valla, Michele and di Volo, Matteo},
	month = aug,
	year = {2024},
	note = {Publisher: Frontiers},
	pages = {1422159},
}

@article{divolo2022,
	title = {Coherent oscillations in balanced neural networks driven by endogenous fluctuations},
	volume = {32},
	issn = {1054-1500},
	url = {https://doi.org/10.1063/5.0075751},
	doi = {10.1063/5.0075751},
	number = {2},
	urldate = {2024-02-10},
	journal = {Chaos: An Interdisciplinary Journal of Nonlinear Science},
	author = {di Volo, Matteo and Segneri, Marco and Goldobin, Denis S. and Politi, Antonio and Torcini, Alessandro},
	month = feb,
	year = {2022},
	pages = {023120},
}

@article{rossler1976,
	title = {An equation for continuous chaos},
	volume = {57},
	issn = {0375-9601},
	url = {https://www.sciencedirect.com/science/article/pii/0375960176901018},
	doi = {10.1016/0375-9601(76)90101-8},
	number = {5},
	urldate = {2025-02-13},
	journal = {Physics Letters A},
	author = {Rössler, O. E.},
	month = jul,
	year = {1976},
	pages = {397--398},
}

@article{pathak2024,
	title = {Biomimetic model of corticostriatal micro-assemblies discovers a neural code},
	author = {Pathak, Anand and Brincat, Scott L. and Organtzidis, Haris and Strey, Helmut H. and Senneff, Sageanne and Antzoulatos, Evan G. and Mujica-Parodi, Lilianne R. and Miller, Earl K. and Granger, Richard},
	journal = {Nature Communications},
	volume = {17},
	number = {1},
	pages = {390},
	year = {2026},
	doi = {10.1038/s41467-025-67076-x},
	url = {https://doi.org/10.1038/s41467-025-67076-x},
	issn = {2041-1723},
}

@article{udrescu2020,
	title = {{AI} {Feynman}: {A} physics-inspired method for symbolic regression},
	volume = {6},
	copyright = {Copyright © 2020 The Authors, some rights reserved; exclusive licensee American Association for the Advancement of Science. No claim to original U.S. Government Works. Distributed under a Creative Commons Attribution NonCommercial License 4.0 (CC BY-NC).. This is an open-access article distributed under the terms of the Creative Commons Attribution-NonCommercial license, which permits use, distribution, and reproduction in any medium, so long as the resultant use is not for commercial advantage and provided the original work is properly cited.},
	issn = {2375-2548},
	shorttitle = {{AI} {Feynman}},
	url = {https://advances.sciencemag.org/content/6/16/eaay2631},
	doi = {10.1126/sciadv.aay2631},
	
	number = {16},
	urldate = {2020-04-23},
	journal = {Science Advances},
	author = {Udrescu, Silviu-Marian and Tegmark, Max},
	month = apr,
	year = {2020},
	note = {Publisher: American Association for the Advancement of Science
Section: Research Article},
	pages = {eaay2631},
}

@article{larter1999,
	title = {A coupled ordinary differential equation lattice model for the simulation of epileptic seizures},
	volume = {9},
	issn = {1054-1500},
	url = {http://aip.scitation.org/doi/10.1063/1.166453},
	doi = {10.1063/1.166453},
	number = {3},
	urldate = {2022-09-29},
	journal = {Chaos: An Interdisciplinary Journal of Nonlinear Science},
	author = {Larter, Raima and Speelman, Brent and Worth, Robert M.},
	month = sep,
	year = {1999},
	note = {Publisher: American Institute of Physics},
	pages = {795--804},
}

@article{abrevaya2024effective,
 author = {Germ{\'a}n Abrevaya and Mahta Ramezanian-Panahi and Jean-Christophe Gagnon-Audet and Pablo Polosecki and Irina Rish and Silvina Ponce Dawson and Guillermo Cecchi and Guillaume Dumas},
 journal = {Transactions on Machine Learning Research},
 url = {https://openreview.net/forum?id=uxNfN2PU1W},
 note = {},
 title = {Effective Latent Differential Equation Models via Attention and Multiple Shooting},
 year = {2024},
}

@article{pecora1990,
	title = {Synchronization in Chaotic Systems},
	volume = {64},
	issn = {0031-9007},
	doi = {10.1103/PhysRevLett.64.821},
	number = {8},
	journal = {Physical Review Letters},
	author = {Pecora, Louis M. and Carroll, Thomas L.},
	month = feb,
	year = {1990},
	pages = {821--824},
}

@book{hastie2009,
	address = {New York, NY},
	edition = {2},
	series = {Springer Series in Statistics},
	title = {The Elements of Statistical Learning: Data Mining, Inference, and Prediction},
	isbn = {978-0-387-84857-0},
	doi = {10.1007/978-0-387-84858-7},
	publisher = {Springer},
	author = {Hastie, Trevor and Tibshirani, Robert and Friedman, Jerome},
	year = {2009},
}

@article{paninski2004,
	title = {Maximum Likelihood Estimation of Cascade Point-Process Neural Encoding Models},
	volume = {15},
	issn = {0954-898X},
	doi = {10.1088/0954-898X/15/4/002},
	number = {4},
	journal = {Network: Computation in Neural Systems},
	author = {Paninski, Liam},
	month = nov,
	year = {2004},
	pages = {243--262},
}

@article{truccolo2005,
	title = {A Point Process Framework for Relating Neural Spiking Activity to Spiking History, Neural Ensemble, and Extrinsic Covariate Effects},
	volume = {93},
	issn = {0022-3077},
	doi = {10.1152/jn.00697.2004},
	number = {2},
	journal = {Journal of Neurophysiology},
	author = {Truccolo, Wilson and Eden, Uri T. and Fellows, Matthew R. and Donoghue, John P. and Brown, Emery N.},
	month = feb,
	year = {2005},
	pages = {1074--1089},
}

@article{jiang2026,
	title = {Ensemble Adaptive Libraries and Inner-Product Sparse Regression for Robust Nonlinear Dynamics Discovery},
	volume = {522},
	issn = {0096-3003},
	url = {https://doi.org/10.1016/j.amc.2025.129839},
	doi = {10.1016/j.amc.2025.129839},
	journal = {Applied Mathematics and Computation},
	author = {Jiang, Yuemei and Niu, Xinzheng and Chen, Hao and Zhu, Jiahui and Xie, Kai and Perc, Matja{\v{z}}},
	month = aug,
	year = {2026},
	pages = {129839},
}

\clearpage

\onecolumngrid
\appendix
\section*{Supplemental Material}

\renewcommand{\thefigure}{S\arabic{figure}}
\renewcommand{\thetable}{S\arabic{table}}
\renewcommand{\theequation}{S\arabic{equation}}
\setcounter{figure}{0}
\setcounter{table}{0}
\setcounter{equation}{0}

\subsection*{Detailed neural-population fitting and symbolic recovery}

To make the neural-population fit fully reproducible, we collect its stages here. The microscopic Izhikevich network supplies the simulated population firing rate $r(t)$ and mean voltage $v(t)$ used as the observed training variables, while the recovery current $w(t)$ and synaptic state $s(t)$ remain latent to the loss. For the symbolic model reported in Eq.~\eqref{eq:ccsymb}, the network is fit at $p_c=0.5$ for 40,000 Adam iterations with $K=0.3$. Short continuation stages then reduce the gain to $K=0.1$ before the neural-network residual is projected onto the STLSQ library. The resulting symbolic equations are evaluated as autonomous dynamics, and a separate simulation tests extrapolation to connection probabilities excluded from training.

\begin{figure}[h!]
    \centering
    \includegraphics[width=0.5\textwidth]{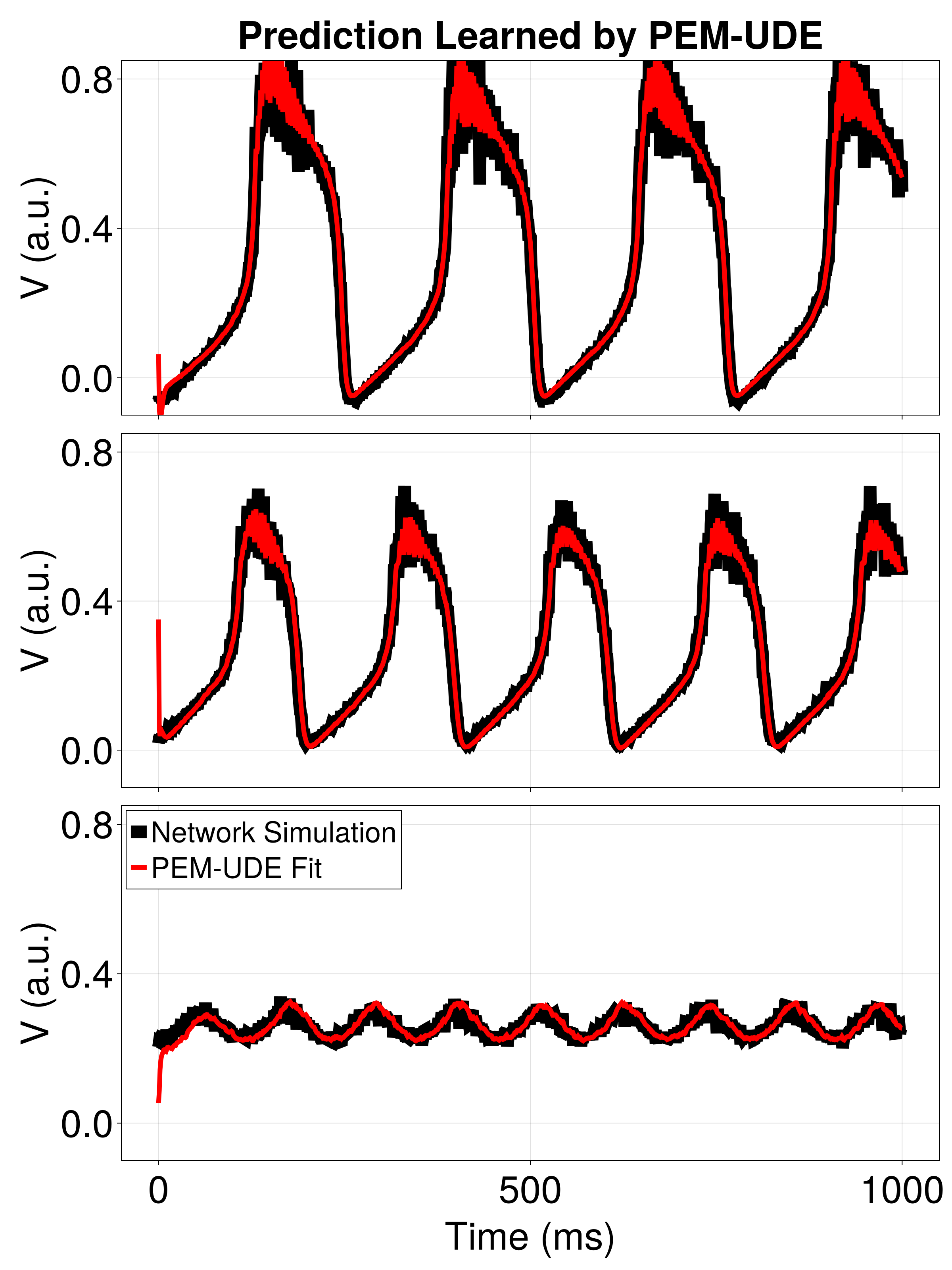}
    \caption{The PEM-UDE learned dynamics, even when constrained during training to samples representing only $p_c = 0.3$ to $p_c = 0.9$, still learn dynamics that generalize down to greater sparsities (e.g., $p_c=0.05$).}
    \label{fig:s4}
\end{figure}

We provide additional notes on the fitting of the PEM-UDE system for the NGNMM at different sparsities, as well as comments on the terms learned by STLSQ in the symbolic form of the system. First, the PEM-UDE approach generalizes to sparsities not in the training data. As an example, in Fig.~\ref{fig:s4} we replot Fig.~\ref{fig:4}e but using a network trained only on $p_c = 0.3$ to $p_c = 0.9$ data. Even though training is limited to intermediate degrees of sparsity, the PEM-UDE system still recovers the dynamics down to 5\% connectivity ($p_c = 0.05$), with a slight frequency shift. While we do not fully exploit this feature in this paper, we note it as a possible direction for further applications of the PEM-UDE approach, which fits well with recent results showing that neural ODEs can generalize beyond bifurcations unseen in the training data \cite{van_tegelen_neural_2025}.

\begin{figure}[h!]
    \centering
    \includegraphics[width=0.85\textwidth]{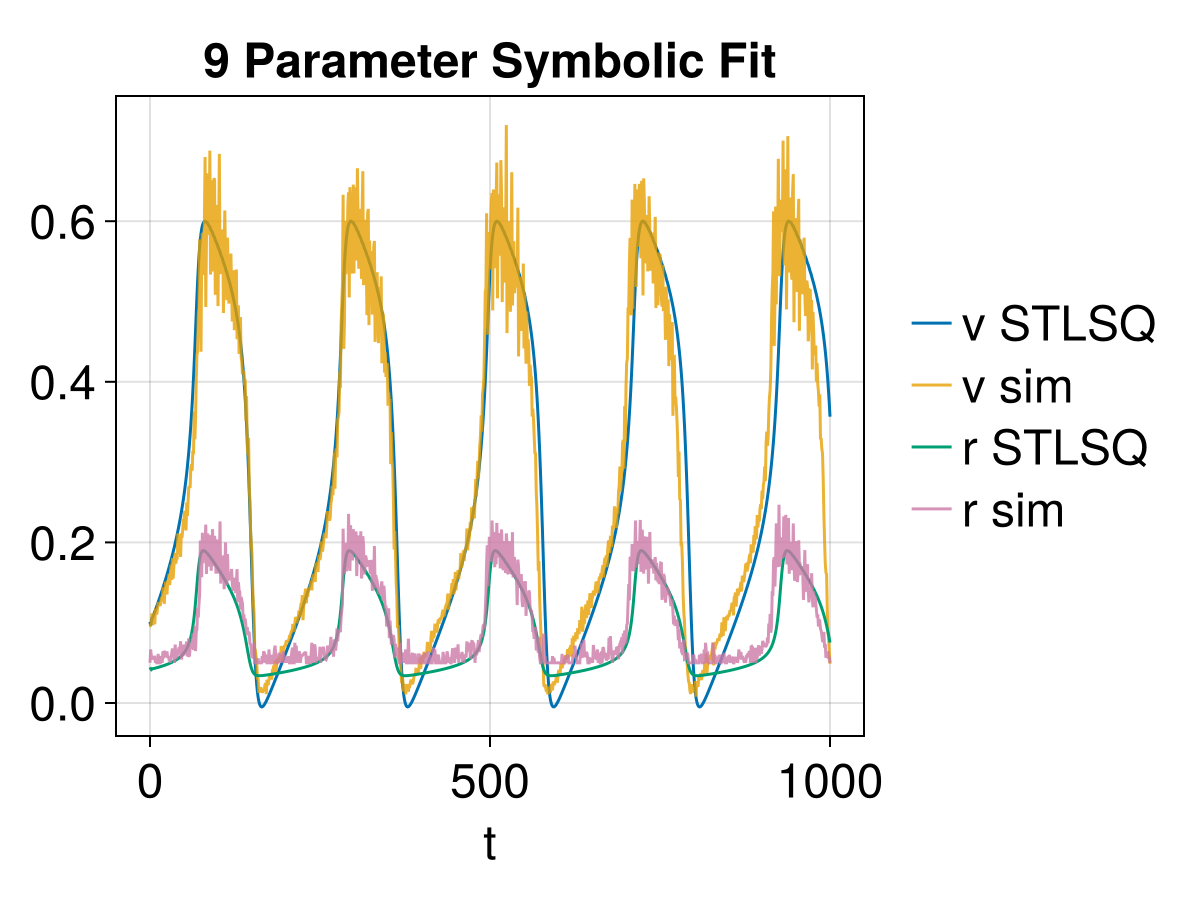}
    \caption{STLSQ-derived symbolic terms presented in Eq.~\eqref{eq:ccsymb} recover dynamics at $p_c=0.5$. See discussion above for how these terms map onto other derivations and biological networks.}
    \label{fig:s5}
\end{figure}

The main advantage of fitting a symbolic form rather than using the UDE black-box dynamics is the ability to analyze the learned terms. This matters most when comparing the new form of the NGNMM with other analytical approaches to mean-field activity of incompletely connected networks of neurons \cite{divolo2018, divolo2022} or networks driven by distributed noisy processes \cite{goldobin2019}. As an illustration, we further consider the dynamics learned from the $p_c = 0.5$ network shown in Fig.~\ref{fig:s5}. Training of the UDE was carried out as follows: the UDE [Eq.~\eqref{eq:cclearn}] is trained using a fixed $p_c = 0.5$, on the simulated firing rate $r$ and average voltage $v$ with $K=0.3$ using the Adam optimizer for 40,000 iterations. We then add shorter training steps during which we slowly reduce $K$ to $0.1$. Below $K=0.1$, the optimization does not improve further. The reason for reducing $K$ is to mitigate any residual influence of the PEM term on the weights of the neural network and to ensure that the UDE reproduces the simulated firing-rate and voltage trajectories with minimal feedback. Applying symbolic regression to the resulting UDE, we find terms similar to those discovered for $\phi_1$--$\phi_6$ in Eq.~\eqref{eq:ccsymb}: a dependence of firing rate on synaptic activity ($\phi_1$) and voltage-dependent synaptic activity ($\phi_3$), along with a linear dependence on $r$ modulated by voltage and synaptic dynamics ($\phi_4$--$\phi_6$). The approach presented here differs most from other derivations in the appearance of a quadratic dependence of voltage on synaptic dynamics ($\phi_7$). This is not present in the simpler network structure but is an emergent feature of supralinear dependence on dendritic integration often seen in biological networks (e.g., NMDA receptor-integrated signaling \cite{poirazi_illuminating_2020}) and used in more detailed neuronal models that focus on dendritic dynamics \cite{abbott_realistic_1991}. Future models should incorporate these dynamics at the population level as well, to provide better ground-truth training data for this approach.

The fit supports three distinct conclusions. First, the PEM-UDE and its symbolic projection reproduce the simulated macroscopic variables in the tested network family. Second, the extrapolation in Fig.~\ref{fig:s4} tests connection probabilities outside the UDE training interval but remains within the same microscopic simulator and parameter family. Third, the individual terms in $f_1$ and $f_2$ are compact closure terms, not uniquely identified biological mechanisms: different residual parameterizations could agree on the sampled trajectories yet differ away from them. Their proposed links to synaptic and dendritic effects should therefore be treated as hypotheses for later analytical or experimental tests.

\subsection*{Supplemental figures}

\begin{figure}[h!]
    \centering
    \includegraphics[width=0.85\textwidth]{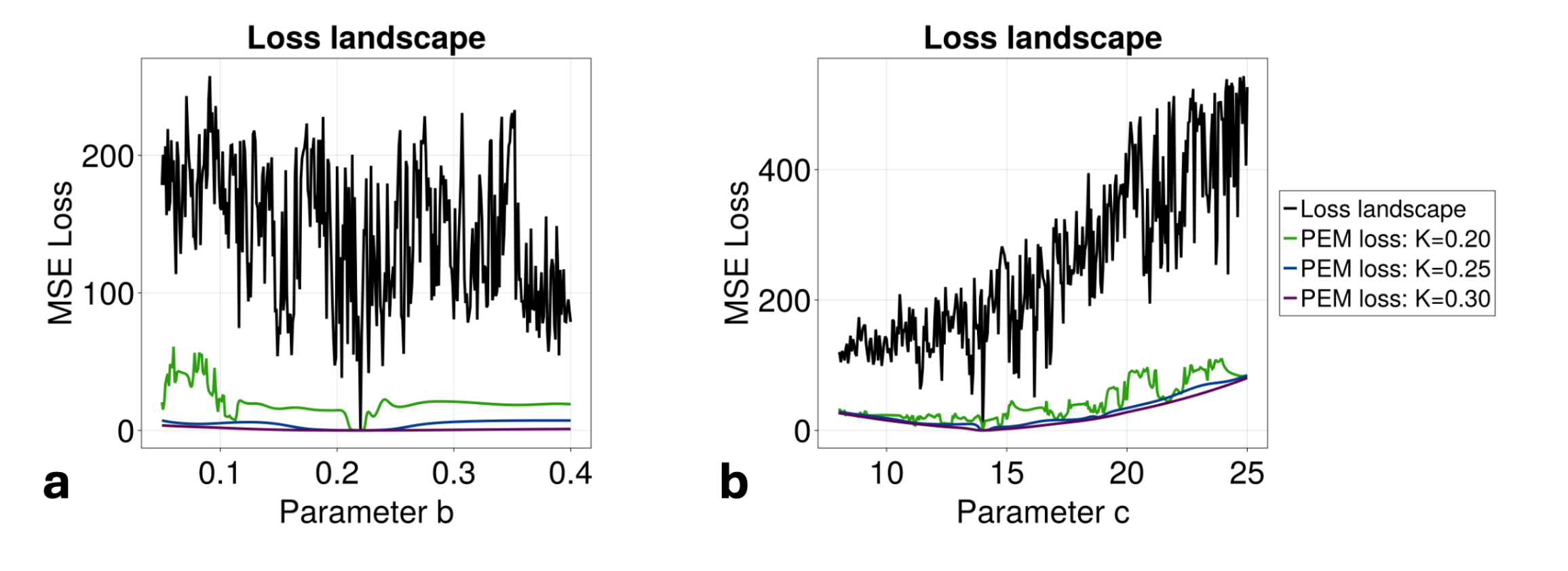}
    \caption{Loss landscapes of the R\"{o}ssler system parameters not fit by the PEM-UDE process as the error correction is applied to the second state. For both parameters $b=0.22$ \textbf{(a)} and $c=14$ \textbf{(b)}, the addition of the correction term to the second state (where neither appears) smooths their loss landscapes but creates a plateau around the true value of the parameter as the correction term is increased. The PEM-UDE fitting method therefore does not alter the landscape in a way that would lead to incorrect dynamics, but the accuracy of the final fit depends on how well the parameters not fit by the PEM-UDE are estimated.}
    \label{fig:s1}
\end{figure}

\begin{figure}[h!]
    \centering
    \includegraphics[width=0.85\textwidth]{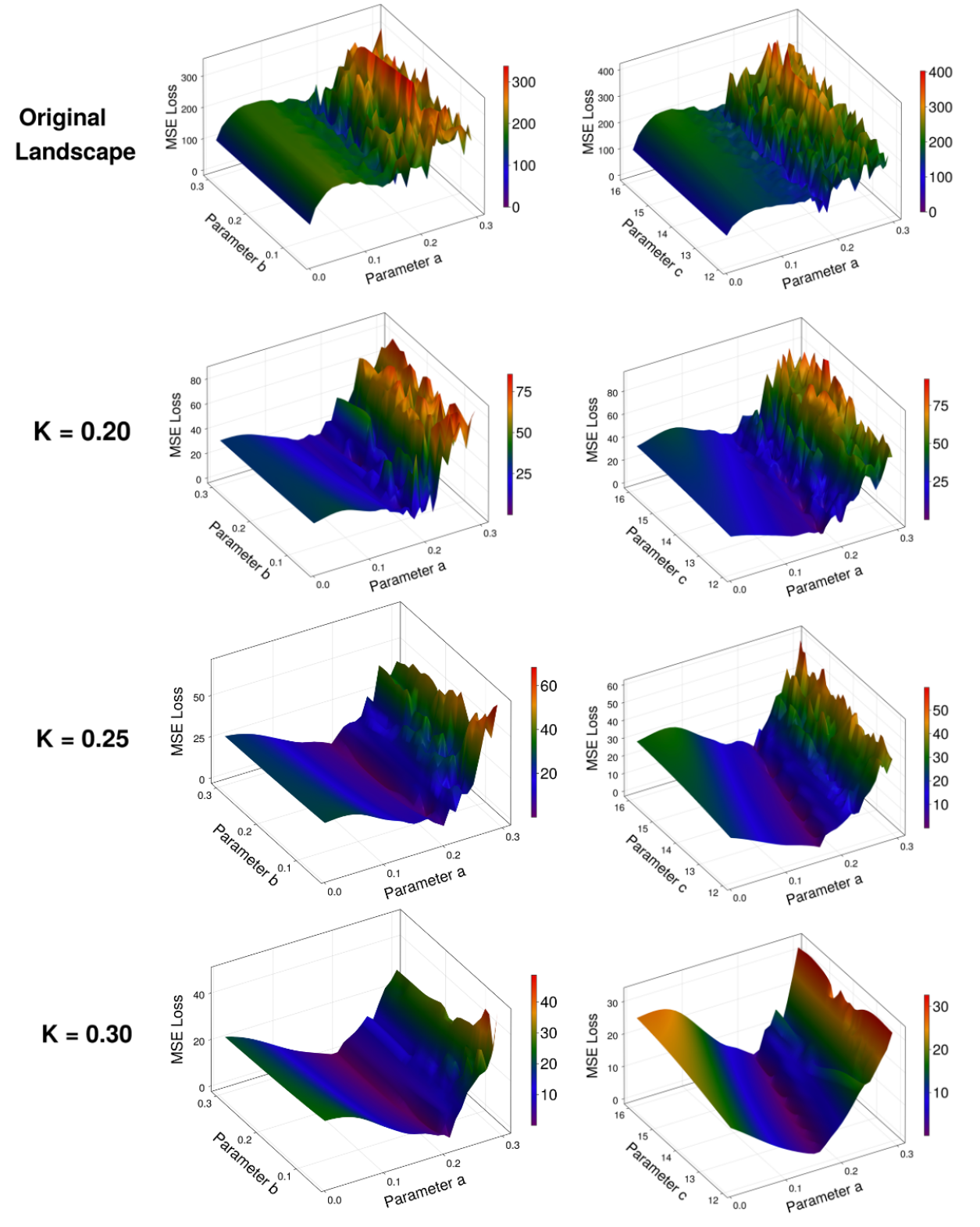}
    \caption{Loss landscapes for the fit ($a$) and unfit ($b$, $c$) parameters when the PEM method is applied to the second state (which contains only $a$). With increasing smoothing $K$, the global minimum becomes more tractable via gradient descent, but a ridge near the minimum requires either knowledge or a good estimate from data of the parameters not being fit to ensure accuracy in the final fit.}
    \label{fig:s2}
\end{figure}

\begin{figure}[h!]
    \centering
    \includegraphics[width=0.85\textwidth]{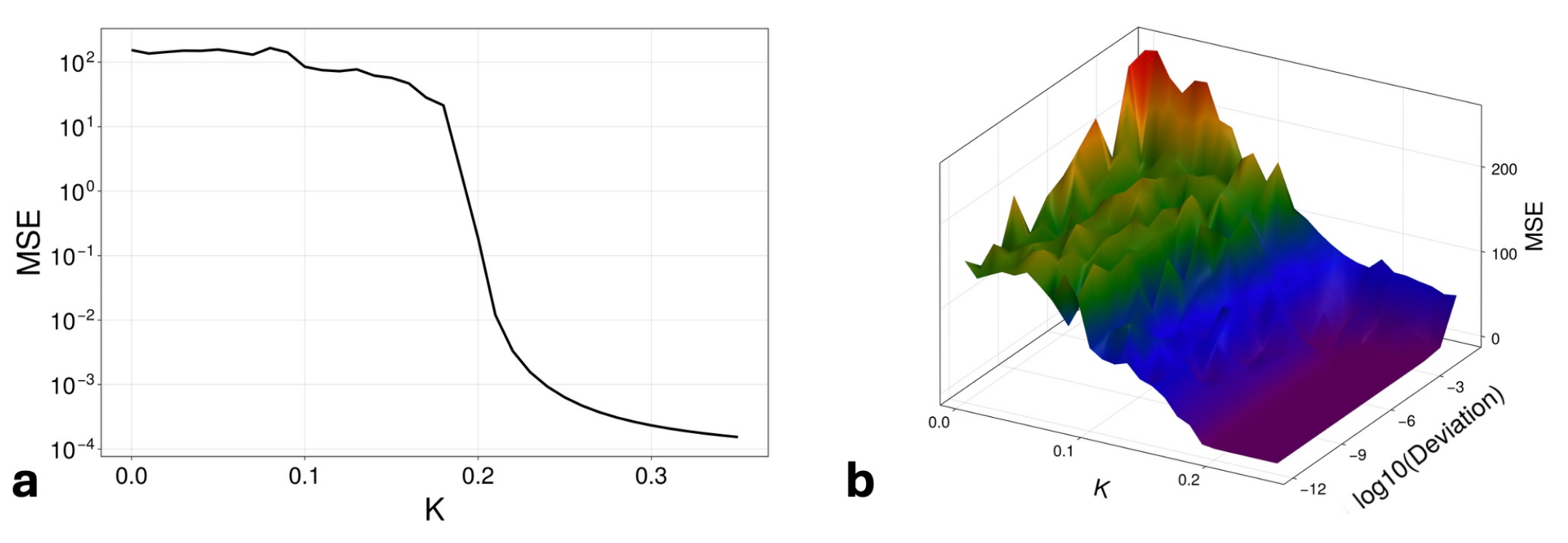}
    \caption{Sufficiently large hyperparameter $K$ for the PEM removes sensitive dependence on initial parameter values. \textbf{(a)} The trajectories of the R\"{o}ssler system diverge for two very close parameter values ($a=0.2$ vs.\ $a=0.200000001$), but this divergence is mitigated by sufficiently large $K$ in the PEM correction term. \textbf{(b)} The correction provides the same aid to fitting for any reasonable deviation from the true parameter (shown for deviations in the range $[10^{-12}, 10^{-1}]$).}
    \label{fig:s3}
\end{figure}

\end{document}